\documentclass{article}
\usepackage[final]{corl_2026}

\usepackage[numbers]{natbib}
\usepackage{multicol}
\usepackage[ruled,linesnumbered]{algorithm2e}
\DontPrintSemicolon

\usepackage{float}
\usepackage{stfloats} 
\usepackage{appendix} 

\usepackage{afterpage}  
\usepackage{float}  
\usepackage{stfloats}  
\usepackage[table]{xcolor}
\usepackage{multirow}
\usepackage{minitoc}
\doparttoc
\usepackage{wrapfig}
\usepackage{caption}
\usepackage{amsfonts}  
\usepackage{amsmath}
\usepackage{booktabs}

\usepackage{graphicx}  
\usepackage{xcolor}

\usepackage{color, soul} 

\definecolor{ourcolor}{HTML}{99e0eb}
\definecolor{ourblue}{HTML}{27a2c3}
\definecolor{tablecolor}{HTML}{ccf2f5}  
\definecolor{tablecolor2}{HTML}{ffcdb4}
\definecolor{citecolor}{HTML}{fe7b5b}
\definecolor{grey}{rgb}{0.9, 0.9, 0.9}
\definecolor{gred}{rgb}{0.859,0.267,0.216}
\definecolor{ggreen}{rgb}{0.059,0.616,0.345}
\definecolor{deepblue}{HTML}{27a2c3}
\definecolor{titleblue}{HTML}{0685f4}
\definecolor{deepred}{HTML}{fe7b5b}

\newcommand{\dd}[2]{$#1\scriptstyle{\pm#2}$}                                            
\newcommand{\ddbf}[2]{\cellcolor{tablecolor}$\mathbf{#1\scriptstyle{\pm#2}}$}           
\newcommand{\cc}[1]{$#1$}                                                               
\newcommand{\ccbf}[1]{\cellcolor{tablecolor}$\mathbf{#1}$}                              

\usepackage{kantlipsum}
\usepackage{xcolor}
\usepackage{etoolbox}
\usepackage{capt-of}
\usepackage{bbm}
\usepackage{amssymb}

\usepackage{enumitem}
\setlist[itemize]{noitemsep, topsep=0pt}

\newcommand{\insertfig}{
    \begin{center}
    \includegraphics[width=\linewidth]{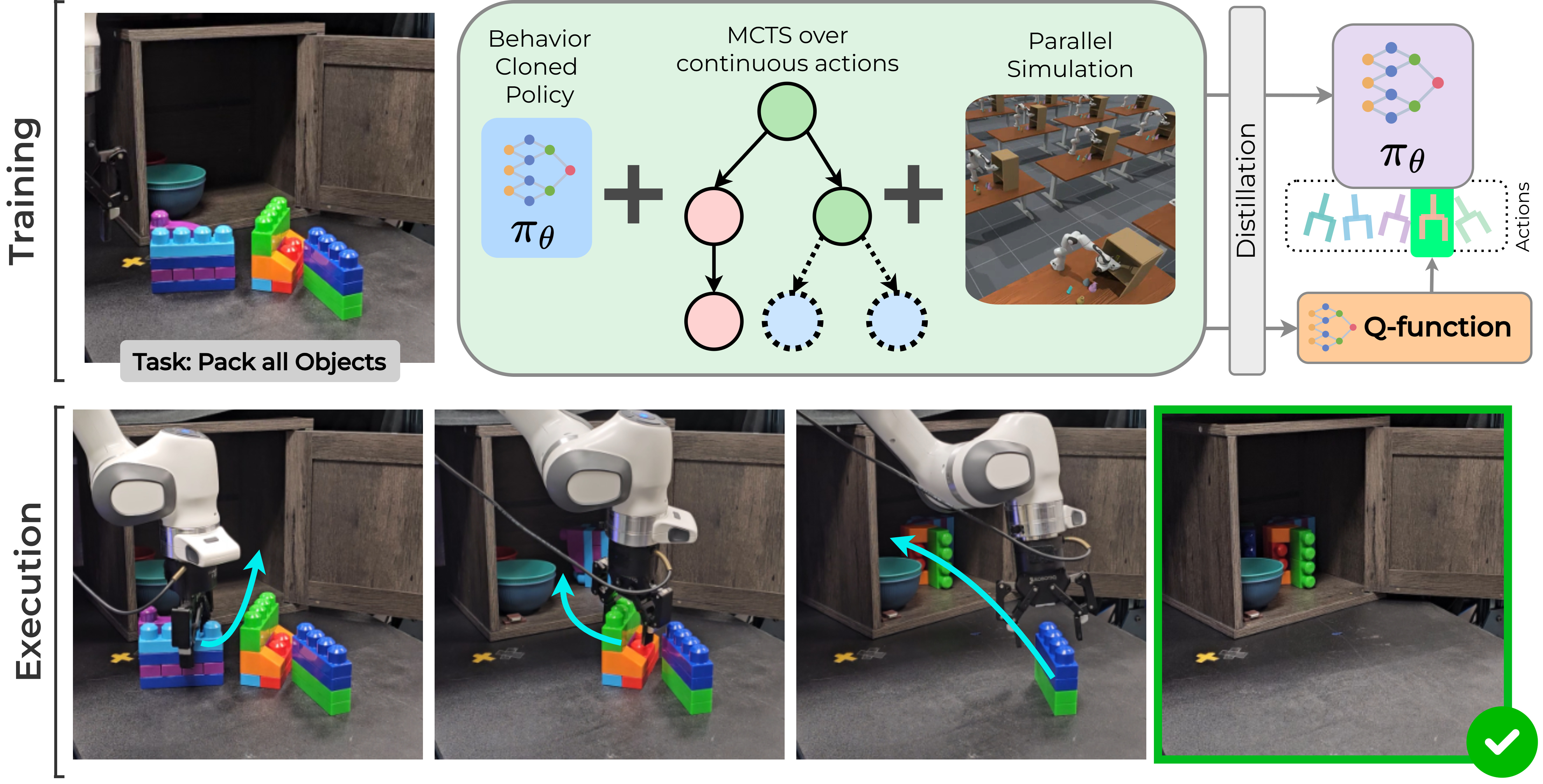}

    \captionof{figure}{\small \textbf{DISCO: \underline{\textbf{Di}}stilled Tree \underline{\textbf{S}}earch over \underline{\textbf{Co}}ntinuous Actions.}
DISCO addresses long-horizon multi-object rearrangement by using progressive-widening Monte Carlo Tree Search over continuous actions. Using an initial behavior-cloned policy to propose short-horizon actions, DISCO expands search trees in parallel GPU simulation to discover long-horizon plans. The search trees are then distilled into a policy and a value function, enabling direct rollout on held-out tasks, such as real-world shelf packing. MCTS and simulation are used only during training. At test time, the distilled model acts directly from point clouds.
}
    \label{fig:teaser}
    \end{center}
}

\makeatletter
\renewcommand{\section}{\@startsection{section}{1}{\z@}%
  {-1.6ex \@plus -0.3ex \@minus -0.2ex}{0.9ex \@plus 0.2ex}{\large\bf\raggedright}}
\renewcommand{\subsection}{\@startsection{subsection}{2}{\z@}%
  {-1.3ex \@plus -0.3ex \@minus -0.2ex}{0.5ex \@plus 0.1ex}{\normalsize\bf\raggedright}}
\makeatother
\begin{document}
\faketableofcontents

\title{Long-Horizon Multi-Object Rearrangement via Distilled Tree Search over Continuous Actions}

\author{
  Kallol Saha$^{1}$, \ Yining Hong$^{2}$, \ Amber Li$^{1}$, \ Yejin Choi$^{2}$, \\[5px]
  \textbf{\ Maxim Likhachev$^{1}$, \ David Held$^{1}$} \\[5px]
  $^1${Robotics Institute, Carnegie Mellon University} \ \ \ \ \ \
  $^2${Stanford University} \\[7px]
}
 

\maketitle

\vspace{-30pt}

\begin{abstract}
Despite recent advances in policy learning, long-horizon manipulation remains difficult because learned policies must avoid compounding errors while preserving future feasibility. While search-based planning can explicitly reason over such future consequences, it quickly becomes expensive in high-dimensional continuous action spaces.
This raises a central question: how can long-horizon reasoning from search be transferred into policies in \textit{continuous action spaces}? To this end, we propose \textbf{DISCO}: \underline{Di}stilled Tree \underline{S}earch over \underline{Co}ntinuous Actions. DISCO uses progressive-widening MCTS in physics simulation to generate long-horizon plans. During search, a learned policy proposes continuous actions, while a learned value function scores actions by their expected progress toward task completion. Using GPU-parallelized simulation, DISCO periodically distills MCTS-generated plans into the policy-value model, improving the action proposals and value estimates used in subsequent searches. Finally, the learned policy-value model can be directly deployed without the need for simulator-based search at test time. Across simulated and real-world rearrangement tasks, DISCO outperforms behavior cloning, reinforcement learning, and search-based baselines.

\textbf{Website:}
\url{https://distilled-search.github.io}
\end{abstract}
\vspace{-10pt}

\vspace{-7pt}

\insertfig

\vspace{-10pt}
\section{Introduction}
\label{sec:intro}
Robot learning has advanced rapidly in recent years, enabling neural network policies to acquire complex behaviors across a wide range of tasks in robot manipulation~\citep{3dfa,3dda,peract,rvt,bridgevla,zhou2023hacman,jiang2024hacmanpp}.
Despite their strong performance on short-horizon behaviors, these methods still struggle with long-horizon manipulation tasks such as multi-object rearrangement~\citep{stilman2007mamo, saha2025spot, huang2024points2plans, wang2022efficient, kee2023sdf, kulshrestha2023structural, wu2022transporters}. Such tasks require reasoning over many sequential decisions, where each action must be locally feasible while preserving future task feasibility. This makes behavior-cloned policies brittle: errors compound over long sequences~\cite{simchowitz2025pitfalls, ross2011dagger}, and short-horizon supervision does not capture how early actions affect later success~\cite{learning2search, learningtrialserrorsreflective, sailor}. For example, shelf packing requires fine-grained pick-and-place actions that are feasible and preserve space for later objects (Figure~\ref{fig:teaser}).
Meanwhile, search-based planning~\cite{m4m_icra2023, em4m_icaps2023, improvedm4m_icra2024, cuTAMP, forwardsearchplanning} provides a natural mechanism for reasoning over future consequences of an action, by evaluating whether it leads to successful downstream configurations. Building on this, the integration of learning with tree search has been powerful in solving long-horizon decision-making problems like Go, chess, and Atari. Most notably, AlphaGo~\cite{alphago, alphago-zero, muzero, anthony2017exit} combines Monte Carlo Tree Search (MCTS)~\cite{banditmcts,mctssurvey} with learned policy and value networks to achieve strong sequential reasoning capabilities. In these systems, search explores the future consequences of actions, while learned policies and value functions guide exploration toward promising regions of the decision space.
However, these domains are highly structured, with discrete actions and exact transition dynamics. In contrast, robots see noisy, partially-occluded observations such as point clouds and operate in continuous action spaces. As a result, naively applying tree search quickly becomes computationally expensive.
This raises a question: \textit{how can the long-horizon reasoning of search be transferred into learned policies for robot manipulation in continuous state-action spaces?}

In this work, we take a step toward bridging this gap with \textbf{DISCO}, a framework for \textbf{DI}stilled tree \textbf{S}earch over \textbf{CO}ntinuous actions. DISCO combines a learned manipulation policy with MCTS using progressive widening to efficiently search over continuous action spaces~\cite{chaslot2008progressive,continuous-uct,mcts-optimization}. The policy proposes continuous grasp-and-place actions conditioned on point clouds, while a learned value function estimates long-term task completion to guide search towards promising regions. To efficiently explore the space of feasible long-horizon plans, we leverage GPU-parallelized simulation for batched node expansion and rollout evaluation~\cite{huang2022parallel,cuTAMP}. Similar to AlphaGo, DISCO distills MCTS-generated plans back into both the policy and the value function, allowing search to progressively discover long-horizon plans. The distillation process also enables DISCO to directly execute long-horizon behaviors
during execution (Figure~\ref{fig:teaser}). Search and simulation are used only during training: DISCO trains one shared policy-value model per task on randomized simulated instances, and at test time this model acts directly from point clouds on unseen instances, without search or a simulator. In this way, DISCO incorporates long-horizon reasoning capabilities into manipulation policies through iterative search-based self-improvement.
We evaluate DISCO on several challenging multi-object rearrangement tasks in simulation and in the real world. We observe that DISCO achieves stronger long-horizon success than behavior cloning, reinforcement learning, sampling-based optimization, and alternative search-based planners. We also show that DISCO benefits from both policy-guided expansion and progressive widening, scales with task complexity, and transfers to real-world manipulation tasks. Our main contributions are:\\[3px]
(1) We show that long-horizon behavior can be bootstrapped purely from short-horizon data (e.g., single-object pick-and-place) by using MCTS with progressive widening to search over continuous state-action spaces. This removes the need for expensive long-horizon demonstrations.\\[3px]
(2) We show that DISCO self-improves both the policy and the value function through iterative distillation rounds, enabling search to progressively generate more long-horizon plans for supervision. \\[3px]
(3) We show that the distilled point-cloud-conditioned policy-value model executes long-horizon tasks by direct rollout, without search or a simulator at test time, scales better than baselines as the number of objects increases, and transfers to held-out simulation tasks and a real robot.


\section{Related Work}
\label{sec:related-work}

\textbf{Combined Learning and Planning.}
Recent Task and Motion Planning (TAMP) methods~\cite{tamp,pddlstream,cuTAMP} incorporate learning for continuous samplers~\cite{skillcomposing,silver2023neurosymbolicskills,chitnis2016guidedsearch,kim2019scorespace}, symbolic abstractions and operators~\cite{silver2021symbolicoperators,silver2023neurosymbolicskills,silver2023predicateinvention,curtis2022abstractions,IVNTR}, skill and parameter models~\cite{liang2022skillmodels,kumar2024practice}, and search heuristics~\cite{valuefunctiontamp,yang2023piginet,khodeir2023lazysearch,forwardsearchplanning}. These methods rely on symbolic representations, whereas DISCO operates directly on point-cloud observations and predicts continuous actions. Skill-chaining methods~\cite{skilldiscovery,lee2022adversarialskillchaining,genskillchaining,mosaic,logicskillprogramming,planning-with-diffusion} compose temporally extended skills or options~\cite{options,option-critic} for long-horizon planning. Rather than learning a skill graph or composing separate skill models, DISCO distills plans into a policy-value model. World-model-based planning methods learn transition models for imagined rollouts or model-predictive control~\cite{ha2018worldmodels, dino-wm, hafner2019planet,hafner2020dreamer,hansen2022tdmpc,li2025worldmodelrobotplanning,khorrambakht2025worldplanner}. In contrast, DISCO does not learn a transition model, and uses physics simulation only during training to generate long-horizon supervision. Other approaches plan from point clouds using object-centric transformations~\cite{saha2025spot}, relational abstractions~\cite{huang2024points2plans, huang2023erdtransformer,pairwiserd}, or neural object descriptors~\cite{simeonov2021long,cheng2024nodtamp}. DISCO instead learns a closed-loop policy-value model and executes without explicit tree search at inference.\\[3px]
\textbf{Plan Distillation and Policy Learning.}
Prior work has distilled planning pipelines into neural policies, including TAMP imitation~\cite{dalal2023optimus} and neural motion-planner distillation~\cite{mpd,saha2024edmp,dalal2024neuralmp,drp}. Related learning-to-search methods~\cite{learning2search,sailor} train policies to reduce compounding errors from imitation learning~\cite{ross2011dagger}. In contrast, DISCO iterates between planning and learning: generated long-horizon plans update the policy-value model, which then guides subsequent plan generation. Recent 3D visuomotor policies map point clouds~\cite{3dfa,3dda,zhou2023hacman,jiang2024hacmanpp}, voxel grids~\cite{peract}, or multi-view features~\cite{rvt,bridgevla,manipgen} directly to robot actions, but are typically trained from fixed-horizon demonstrations. DISCO instead self-generates long-horizon plans to train a closed-loop model for execution.
\\[3px]
\textbf{Monte-Carlo Tree Search.}
For continuous high dimensional action spaces, prior work extends MCTS using progressive widening~\cite{chaslot2008progressive,continuous-uct,mcts-optimization}, learned action sampling~\cite{kim2017guidingsearchcontinuousstateaction}, policy priors~\cite{pa-mcts,sampledmuzero}, state abstraction~\cite{iterativelyrefiningmcts}, or prioritized branching~\cite{improvedpomdptreesearch,monte-carlo-pomdp,onlinePOMDP}. In robotics, MCTS has been used for nonprehensile rearrangement~\cite{song2020multi}, clutter retrieval~\cite{huang2022visual,huang2022interleaving}, batched long-horizon planning~\cite{huang2022parallel}, contact planning~\cite{zhu2023efficient}, confined-space rearrangement and occluded-object grasping~\cite{ren2024multistage,kang2022grasp}, multi-goal task planning~\cite{pfeiffer2023monte}, and world-model-based planning~\cite{khorrambakht2025worldplanner,huang2025forgetree}. AlphaGo-style systems further show that MCTS can be coupled with policy-value learning in a self-improving loop~\cite{alphago,alphago-zero,muzero,anthony2017exit}, and dual policy iteration~\cite{sun2018dpi} alternates between a planning policy and a learned reactive policy. Neural Tree Expansion~\cite{riviere2021nte} guides MCTS with learned policy-value models for continuous multi-robot planning, but runs MCTS online at test time, whereas DISCO runs MCTS only during training. DISCO extends progressive-widening MCTS over continuous actions into robot manipulation problems, using a point-cloud conditioned policy-value model that is iteratively improved during search.

\section{Problem Statement and Assumptions}
\label{sec:problem}
We consider a distribution over long-horizon, multi-object robot manipulation task instances $\mathcal{T}$. Each instance
consists of a discrete set of $N_o$ objects. 
The state $s\in\mathcal{S}\subseteq\mathbb{R}^{{N_o}\times 6}$ is represented by a 6D pose for
each object. The initial state $s_0\in\mathcal{S}$ varies across task instances. A goal state
$s_K$ should satisfy $\mathcal{G}(s_K)=1$, where
$\mathcal{G}:\mathcal{S}\rightarrow\{0,1\}$ checks
whether the object poses complete the task instance, such as being placed in
a shelf or stacked correctly. At test time, we only have access to a point cloud $\mathbf{o}$
, with each point assigned a segmentation label for an object. 
We operate in a \textit{continuous action space} $\mathcal{A}$, where an action $\mathbf{a}\in\mathcal{A}$ is represented by target end-effector poses.
We use motion planning~\cite{curobo} to convert the poses into a trajectory. During training, but not during inference, we assume access to a simulator
$f_{\mathrm{sim}}$~\cite{maniskill} that executes an action to induce the next state $s'$. 
We define a valid plan $\rho$ as a sequence of states and actions, $\rho = [s_0,\mathbf{a}_0,s_1,\ldots,\mathbf{a}_{K-1},s_K]$,
satisfying the goal condition $\mathcal{G}(s_K)=1$ and the simulator transition constraint
$s_{k+1}=f_{\mathrm{sim}}(s_k,\mathbf{a}_k)$.

We assume access to a dataset
$\mathcal{D}=\{(\mathbf{o}_k,\mathbf{a}_k)\}_{k=1}^{N_d}$ of observation-action tuples that demonstrate subtasks (e.g. Pick or Place). Importantly, we do not assume access to any demonstrations that solve the long-horizon task.
Only in simulation, we define a task progress $\kappa:\mathcal{S}\rightarrow[0,1]$ (Appendix~\ref{app:task_setup}), and a progress reward $r(s,\mathbf{a},s')=\kappa(s')-\kappa(s)$ for executing action $\mathbf{a}$ from state $s$ and reaching state $s'$. The objective is to train a model that maximizes the probability of reaching a goal state on any instance of the long-horizon task $\mathcal{T}$. Appendix~\ref{app:markov} formalizes each task instance as a goal-conditioned Markov decision process and states this objective.
\section{DISCO: Distilled Search over Continuous Actions}
\label{sec:method}
We present a framework for distilling a closed-loop policy $\pi_{\theta}$ and value-function $Q_{\phi}$ to solve unseen instances of a long-horizon task $\mathcal{T}$. We use MCTS with the learned models to self-generate long-horizon plans for supervision on simulated task instances $\mathcal{T}_{\mathrm{sim}} \subset \mathcal{T}$. At test time, the model maps each observation $\mathbf{o}_k$ to an action $\mathbf{a}_k$ without any explicit search or simulator calls. The central challenge here is searching over a continuous action space and selecting actions that remain locally feasible while making progress toward long-horizon task completion, such as leaving space for future objects in shelf packing. We initialize the policy from single-object pick-and-place demonstrations. During search, the policy proposes actions for expansion, 
the value function guides rollouts, and both networks are iteratively updated in intervals set by a fixed number of MCTS iterations.

\subsection{Monte-Carlo Tree Search over Continuous Actions}
\label{sec:mcts}
We define a directed acyclic search graph $G = \langle V, E\rangle$, with $N_v$ vertices and edges $V = [\mathbf{v}_0,\ldots,\mathbf{v}_{N_v}]$ and $E = [\mathbf{e}_1,\ldots,\mathbf{e}_{N_v}]$ respectively. Each vertex $\mathbf{v}_i \in V$ stores a simulation state
$s_i \in \mathcal{S}$, a reward $r_i$, a visit count
$N(\mathbf{v}_i)$, and a value $Q(\mathbf{v}_i)$ estimating its utility toward
task success. 
The search graph
is a rooted tree: every non-root vertex is connected to its parent by a unique incoming edge
$\mathbf{e}_i \in E$, has a unique
$\mathbf{v}_p =\mathrm{parent}(\mathbf{v}_i)$ and
$\mathrm{children}(\mathbf{v}_i) =
[\mathbf{v}_{c_1},\mathbf{v}_{c_2}\ldots]$. The reward for an edge is computed as $r_i=r(s_p,\mathbf{a}_i,s_i)$, where
$\mathbf{v}_p=\mathrm{parent}(\mathbf{v}_i)$ and $s_p$ is the parent vertex's state. 
Each edge stores an action parameter
$\mathbf{a}_i$.
\begin{figure}[t]
    \centering
    \includegraphics[width=\linewidth]{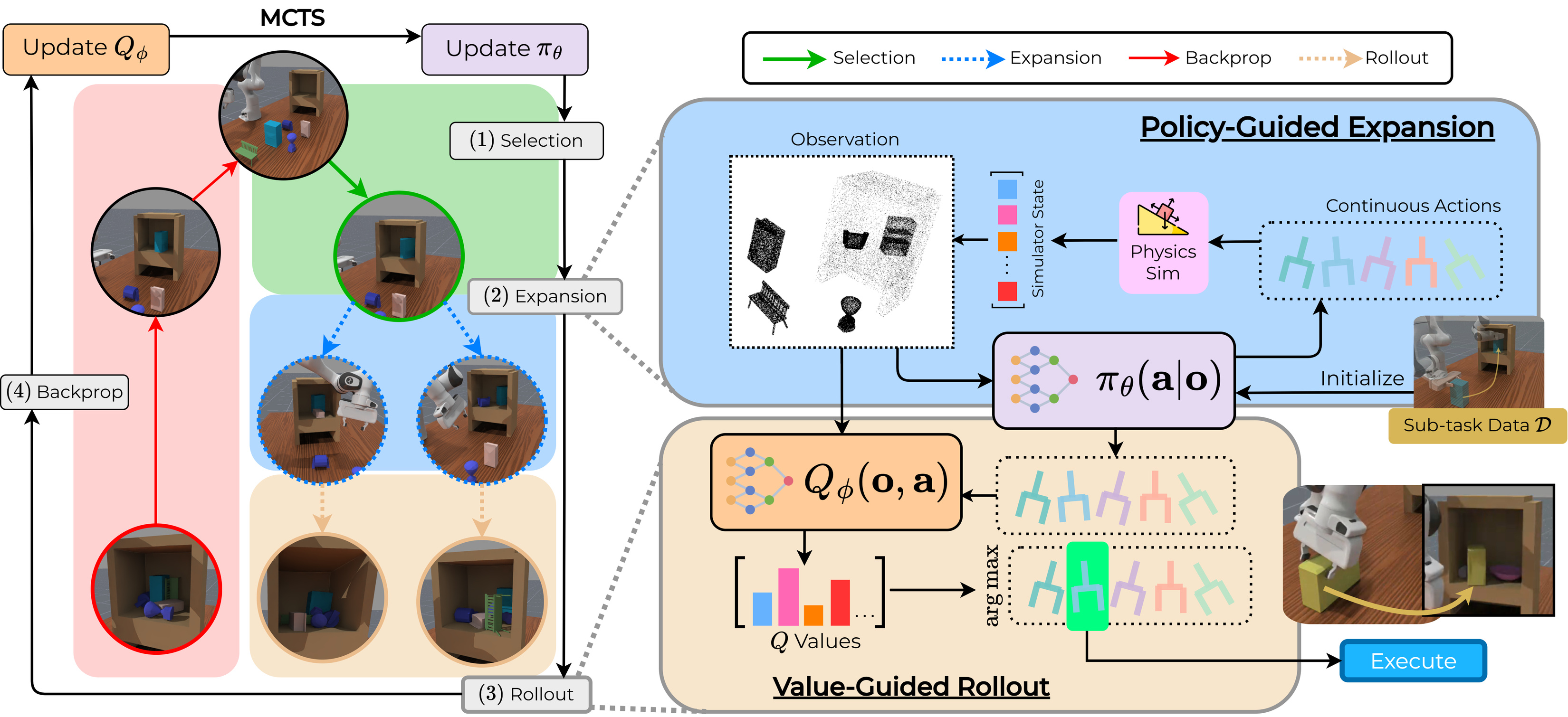}
    \caption{\small \textbf{MCTS with DISCO}. We embed a policy $\pi_{\theta}$ and a value function $Q_{\phi}$ into the MCTS loop. The policy proposes continuous actions for tree expansion, while the value function guides rollout action selection toward terminal states. Periodically, both models are updated from newly generated search trees, enabling direct policy-value deployment at test time.}
    \label{fig:method}
    \vspace{-12pt}
\end{figure}

\textbf{Vertex Expansion.} We use a policy $\pi_{\theta}(\mathbf{a} | \mathbf{o})$ for expanding vertices during search (Figure~\ref{fig:method}). We pre-train the policy by behavior cloning from the dataset $\mathcal{D}$ (Appendix~\ref{app:pretrain_dataset}). We use a modified version of 3D FlowMatch Actor (3DFA)~\cite{3dfa} for the policy architecture (Appendix~\ref{app:policy_arch}).
For expanding a vertex $\mathbf{v}_i$, we sample actions from the policy $\mathbf{a}_{i+1} \sim \pi_{\theta}(\mathbf{a}_{i+1} | \mathbf{o}_i)$, using the observation $\mathbf{o}_i$ from the vertex. After executing the action through the simulator, we obtain the new physical state $s_{i+1}=f_{\mathrm{sim}}(s_i,\mathbf{a}_{i+1})$ of the system.
We then add a new vertex $\mathbf{v}_{i+1}$ with state $s_{i+1}$ and a new edge $\mathbf{e}_{i+1} = \left\langle \mathbf{a}_{i+1}, \mathbf{v}_{i+1}\right\rangle$ to the graph, $G \gets G \cup \langle \mathbf{v}_{i+1}, \mathbf{e}_{i+1} \rangle$.


\setlength{\columnsep}{15pt} 

\begin{wrapfigure}{R}{0.47\linewidth}
\vspace{-4pt}
\begin{algorithm}[H]
\footnotesize
\caption{DISCO}
\label{alg:disco}
\KwIn{$\pi_\theta,Q_\phi,f_{\mathrm{sim}},r,\mathcal{G},(k,\alpha)$}
\KwOut{Updated $\pi_\theta$, updated $Q_\phi$}

Initialize $G=\langle V,E\rangle$ with root $\mathbf{v}_0$, $V_{\mathrm{succ}}\gets\emptyset$\;

\While{search is running}{
    $\mathbf{v}\gets$ \textsc{UCBSelect}$(G)$\;

    \If{$|\mathrm{children}(\mathbf{v})| < kN(\mathbf{v})^\alpha$}{
        $G\gets$ \textsc{Expand}$(G,\mathbf{v},\pi_\theta,f_{\mathrm{sim}})$\;
    }

    $R\gets$ \textsc{Rollout}$(\mathbf{v},\pi_\theta,Q_\phi,r,\mathcal{G})$\;
    $G\gets$ \textsc{Backpropagate}$(G,\mathbf{v},R)$\;

    $V_{\mathrm{succ}}\gets V_{\mathrm{succ}}\cup
    \{\mathbf{v}\in V\mid \mathcal{G}(s_{\mathbf{v}})=1\}$\;

    \If{distillation interval reached}{
        $(\pi_\theta,Q_\phi)\gets$ \textsc{Distill}$(G,V_{\mathrm{succ}})$\;
        Reset $G$ to root $\mathbf{v}_0$, $V_{\mathrm{succ}}\gets\emptyset$\;
    }
}

\Return{$\pi_\theta,Q_\phi$}\;
\end{algorithm}
\vspace{-6pt}
\end{wrapfigure}

\textbf{Progressive Widening.} To efficiently search over the continuous action space  $\mathcal{A}$, we use MCTS with progressive-widening~\cite{chaslot2008progressive,continuous-uct} to gradually increase the set of sampled actions as a vertex receives more visits. This is formalized by expanding a vertex $\mathbf{v}_i$  only when the following condition is met: $|\mathrm{children}(\mathbf{v}_i)| < k\,N(\mathbf{v}_i)^\alpha$, where $k$ and $\alpha$ control the widening rate. This allows adding more children for vertices that are visited more, and constrains expanding new children for less visited vertices of the tree. Since more-visited vertices also have higher Q-values, progressive widening focuses tree expansion in more promising regions that are likely to reach the goal, thus suitable for when we want to efficiently find more plans. Algorithm (\ref{alg:disco}) shows a pseudo-code of our method. Details and hyperparameters are given in Appendices~\ref{app:algorithm_functions} and~\ref{app:mcts_hyperparameters}.
\subsection{Value-Guided Rollout}
After expansion, MCTS estimates the long-horizon utility of the newly added vertex
by rolling out from its state. The rollout is a continuation of the same tree descent rather than a separate rollout policy: at each rollout vertex $\mathbf{v}$ with observation $\mathbf{o}$, we expand $\mathbf{v}$ with a batch of $B$ actions $\{\mathbf{a}^{(b)}\}_{b=1}^{B} \sim \pi_{\theta}(\mathbf{a}|\mathbf{o})$, add every feasible child $\mathbf{v}^{(b)}$ to the search graph as in Section~\ref{sec:mcts}, and continue from the child selected by the UCB rule:
$\mathbf{v}^{*}
=
\arg\max_{\mathbf{v}^{(b)}}
\left[
Q(\mathbf{v}^{(b)})
+
C\sqrt{\frac{\log\left(N(\mathbf{v})+1\right)}{N(\mathbf{v}^{(b)})+1}}
\right],$\\[7px]
where $C$ is an exploration constant that balances exploration and exploitation. A visited child is scored by its backed-up value, while an unvisited child is scored by a prior. In the first round of search, this prior is the progress reward $r$ of its incoming edge. In subsequent rounds, we initialize $Q(\mathbf{v}^{(b)})\gets Q_{\phi}(\mathbf{o},\mathbf{a}^{(b)})$ for value-guided rollouts (Figure~\ref{fig:method}, Appendix~\ref{app:value_prior}). $Q_{\phi}(\mathbf{o},\mathbf{a})$ is a point-cloud and action-conditioned value function that predicts the probability of an action leading to task success (Section~\ref{sec:distill}, architecture in Appendix~\ref{app:value_arch}), trained using the resulting trees as supervision. During value-guided rollouts, the selection rule reduces to executing the highest-scoring sampled action under $Q_{\phi}$. This is an improvement over sampling directly from $\pi_\theta$ (Appendix~\ref{app:success_probability_objective}).
The rollout stops at a terminal state $s_T$ (corresponding to terminal vertex $\mathbf{v}_T$), either because the goal condition $\mathcal{G}(s_T)=1$ is satisfied, the rollout horizon is exhausted, or no feasible action remains.

We compute the return of the rollout as the cumulative progress reward
along the path from root vertex $\mathbf{v}_0$ to terminal vertex $\mathbf{v}_T$, which telescopes to $R=\kappa(s_T)-\kappa(s_0)$.
This return is backpropagated up along the selected path through the tree, including the
newly expanded vertex and all rollout vertices added to the graph. For each
visited vertex, we increment its visit count and update its value estimate using
the running average of returns.
The planner maintains a set of successful terminal vertices
$V_{\mathrm{succ}}=\{\mathbf{v}_j\in V \mid \mathcal{G}(s_j)=1\}$.
For any successful terminal vertex $\mathbf{v}_j\in V_{\mathrm{succ}}$, we
recover the corresponding plan $\rho$ by following parent pointers from
$\mathbf{v}_j$ back to the root $\mathbf{v}_0$.

\textbf{GPU Parallelized Search.} We run several descents of the same tree concurrently and batch their expansions, so a batch consists of state-action pairs $\{(s^{(b)},\mathbf{a}^{(b)})\}_{b=1}^{B}$ whose start states $s^{(b)}$ can belong to different vertices. The actions are
passed through a GPU parallelized motion planner~\cite{curobo} to produce trajectories, which are executed by parallelized GPU simulation~\cite{maniskill}, with each environment reset to its own start state, to obtain a batch of next states
$s'^{(b)}=f_{\mathrm{sim}}(s^{(b)},\mathbf{a}^{(b)})$. All new state-action pairs are scored by the value function in a single batch. Each new state
instantiates a new edge and vertex, adding multiple edges and vertices to
the search graph in parallel.
This speeds up the search process, as compared to evaluating each action
sequentially.
\subsection{Iterative Distillation}
\label{sec:distill}
Our goal is to iteratively update the policy $\pi_\theta(\mathbf{a}_k|\mathbf{o}_k)$ and value function $Q_{\phi}(\mathbf{o}_k,\mathbf{a}_k)$ every few MCTS iterations, so that they improve the likelihood of discovering long-horizon plans in subsequent search iterations and can be directly rolled out on instances of the long-horizon task set $\mathcal{T}$. We aim to do this by distilling the best performance from the expanded search graphs over the simulated task instances $\mathcal{T}_\mathrm{sim}$. For every non-root vertex $\mathbf{v}_i$, we define a value target using the fraction $y_i$ of successful leaf vertices among the visited descendants of $\mathbf{v}_i$, where $\mathrm{leaves}(\mathbf{v}_i)$ denotes the visited descendants of $\mathbf{v}_i$ with no visited feasible children, and train the value function on these targets with a weighted binary cross-entropy loss $\mathcal{L}_V$:
\begin{equation}
\label{eq:value_distill_loss}
\begin{aligned}
    y_i &=
    \frac{
    \sum_{\mathbf{v}_\ell \in \mathrm{leaves}(\mathbf{v}_i)}
    \mathbbm{1}_{\mathcal{G}(s_\ell)}
    }{
    |\mathrm{leaves}(\mathbf{v}_i)|
    },
\end{aligned}
\qquad
\begin{aligned}
    \mathcal{L}_V(\phi)
    &=
    \frac{
    \sum_{i\in\mathcal{B}}
    w_i\,
    \mathrm{BCE}
    \left(
    (Q_\phi(\mathbf{o}_{p_i},\mathbf{a}_i)), y_i
    \right)
    }{
    \sum_{i\in\mathcal{B}} w_i
    } .
\end{aligned}
\end{equation}
This target estimates the probability that the subtree rooted at
$\mathbf{v}_i$ leads to task success, i.e., it is a Monte-Carlo estimate of the success probability that $Q_\phi$ approximates (Appendix~\ref{app:success_probability_objective}). Since targets computed over more leaves are
more reliable, we weigh each target by the number of leaves in its subtree using
$w_i=\log(1+|\mathrm{leaves}(\mathbf{v}_i)|)$. Here, $\mathcal{B}$ is a minibatch of non-root
vertices, $\mathbf{o}_{p_i}$ is the observation at the parent vertex
of $\mathbf{v}_i$, and $\mathbf{a}_i$ is the feasible action stored in its incoming edge
$\mathbf{e}_i$.

The policy is trained from every expanded edge that produced a feasible new
vertex. Each such edge provides supervision $(\mathbf{o}_{p_i},\mathbf{a}_i)$. Training on all feasible edges, rather than only on edges of successful plans, removes probability mass from infeasible actions while keeping the proposals diverse, and leaves ranking them to $Q_\phi$ (Appendix~\ref{app:policy_target}). We train the policy
 using the flow-matching loss defined in 3DFA~\cite{3dfa}.
The sub-task dataset $\mathcal{D}$ is used only for pre-training and is not mixed back in during distillation.
We perform this distillation every $N_{\mathrm{distill}}$ MCTS iterations, and
then continue search by replacing the planner's policy and value function with
the updated $\pi_\theta$ and $Q_\phi$.

\label{sec:methods}

\section{Experiments}
\label{sec:results}
\begin{figure}[t]
    \centering
    \includegraphics[width=\linewidth]{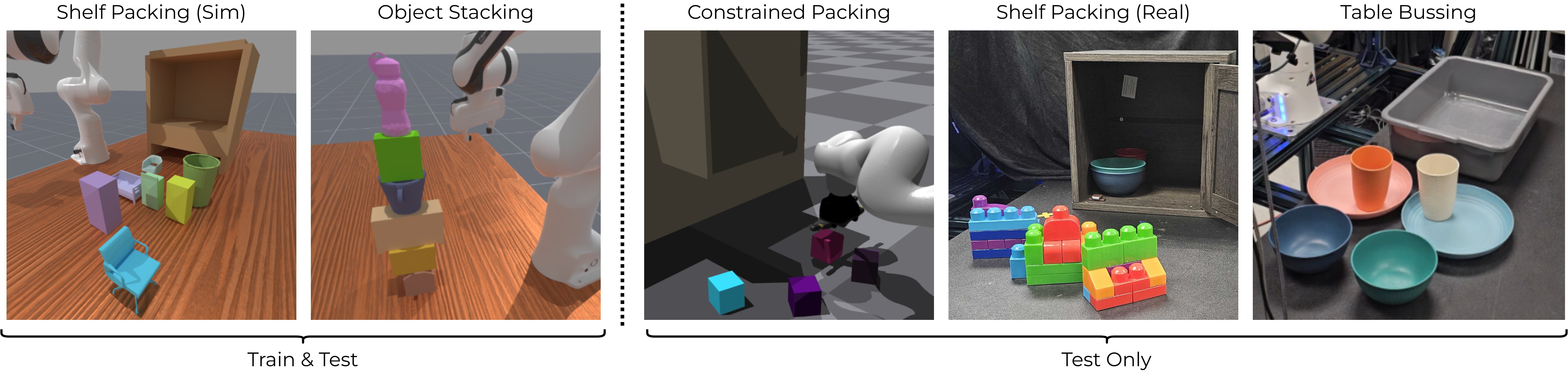}
    \caption{\small \textbf{Task Setup}: Representative states from the tasks used in our evaluation. In Shelf Packing and Object Stacking, the robot must fit objects into a shelf and construct stable stacks, respectively. Constrained Packing requires objects to be placed inside a bounded region without collisions. We also evaluate on a real-world version of Shelf Packing along with a Table Bussing task where dinnerware must be placed upright in a tray.}
    \label{fig:task_setup}
    \vspace{-10px}
\end{figure}

Our experiments seek to answer the following key research questions: \\[3px]
\textbf{RQ1:} Does DISCO make better long-horizon task progress 
than other search-based planners, reinforcement learning (RL), and sampling-based planners?\\
\textbf{RQ2:} How important is enabling search over continuous actions by integrating progressive-widening with a learned policy prior and value function in MCTS?\\
\textbf{RQ3:} Does DISCO scale better than the baselines as task complexity (e.g., object count) increases?\\
\textbf{RQ4:} Does distilling self-generated MCTS supervision back into the policy-value model enable it to solve long-horizon tasks in the real world where a simulator is not available during inference?\\
\textbf{RQ5:} What does iterative distillation learn beyond the single-step pre-training data? 
\subsection{Datasets and Tasks}
We evaluate DISCO on five tasks (Figure~\ref{fig:task_setup}): Shelf Packing, Object Stacking, and Constrained Packing in simulation, and Shelf Packing and Table Bussing in the real world (Appendix~\ref{app:task_setup}).

We implement our simulation tasks in ManiSkill3~\cite{maniskill}. 
For all tasks, observations are point clouds captured from depth cameras, and actions are parameterized by a grasp pose and a placement pose. Each action is executed through cuRobo-based motion planning~\cite{curobo}. We use one shared policy and value function trained on 50 randomized instances per task. We evaluate the models by rolling out on 15 held-out task instances with unseen object layouts. Scenes use ACRONYM~\cite{acronym} object meshes placed randomly on a table. We perform real-world experiments using a Franka Panda robot with a Robotiq gripper. We train and evaluate our models on one RTX 4090 GPU. For shelf packing, search generates 263,650 vertices, of which 11,015 are expanded, in 47.7 hours. Search trees store only simulator states, not point clouds, using 95 MB per tree.

\subsection{Evaluation}


\begin{figure}[t]
\centering
\begin{minipage}{\linewidth}
\centering

\includegraphics[width=\linewidth]{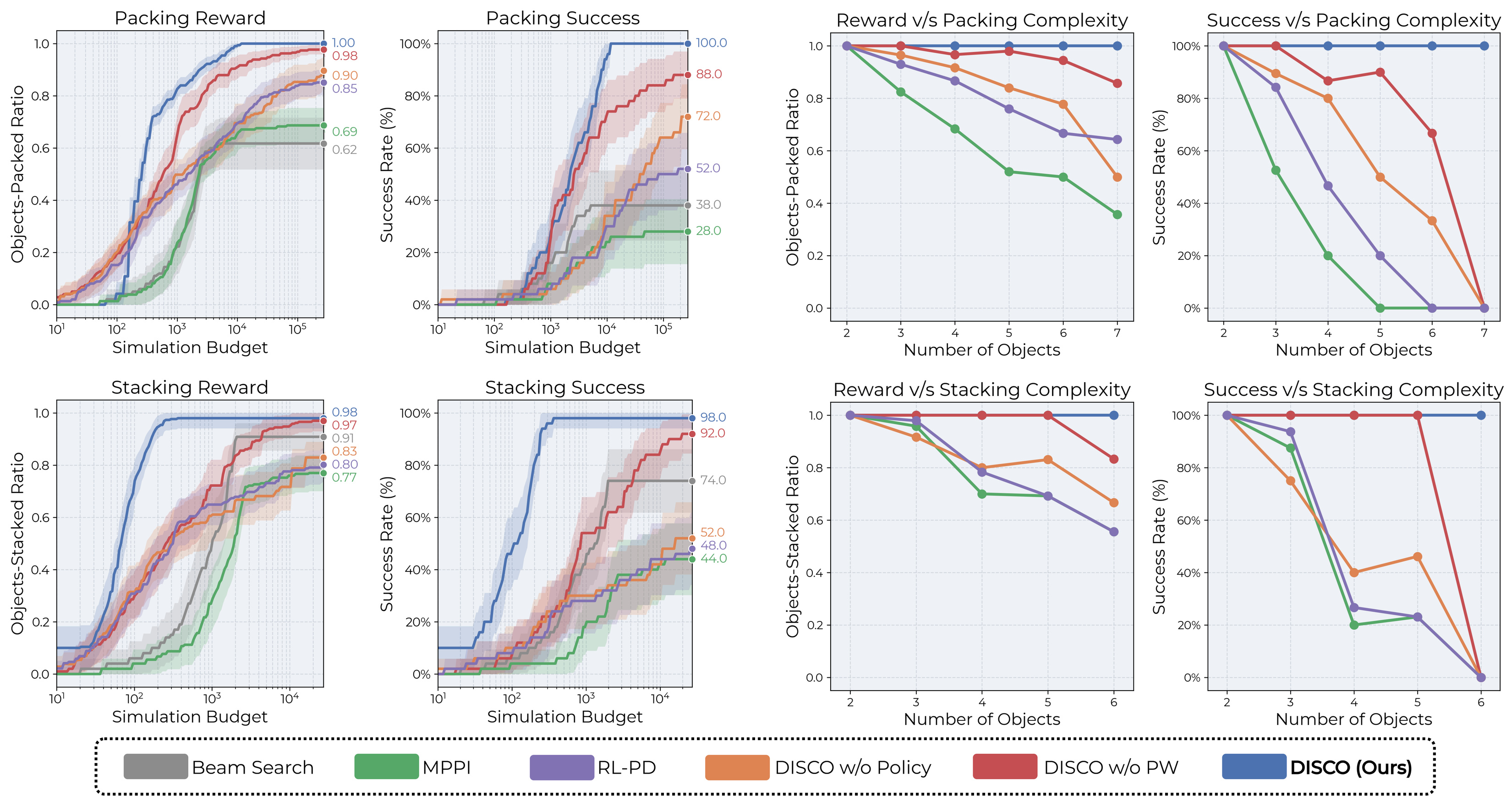}
\captionof{figure}{\small \textbf{Simulation Results.} DISCO improves reward and success with increasing simulation budget on shelf packing and object stacking, and remains more robust as task complexity increases. The simulation budget counts simulator calls made by search during offline training, not at test time.}
\label{fig:results}

\vspace{8pt}

\resizebox{\textwidth}{!}{%
\begin{tabular}{l|ccc|ccc}
\toprule
& \multicolumn{3}{c|}{\textbf{Shelf Packing (Sim) Metrics}}
& \multicolumn{3}{c}{\textbf{Object Stacking Metrics}} \\[5px]
Algorithm
& Reward ($\uparrow$)
& Success (\%) ($\uparrow$)
& Train Plans ($\uparrow$)
& Reward ($\uparrow$)
& Success (\%) ($\uparrow$)
& Train Plans ($\uparrow$) \\
\midrule

Beam Search 
& \dd{24.3}{8.0} 
& \dd{6.7}{5.1} 
& \dd{4.26}{1.14}
& \dd{79.9}{4.2} 
& \dd{33.3}{13.6} 
& \dd{5.84}{1.00} \\

MPPI 
& \dd{36.7}{7.8} 
& \dd{13.3}{7.2} 
& \dd{12.20}{3.25}
& \dd{60.7}{6.2} 
& \dd{13.3}{12.2} 
& \dd{3.32}{0.50} \\

RL-PD 
& \dd{71.0}{4.8} 
& \dd{20.0}{13.7} 
& \dd{138.52}{40.81}
& \dd{64.5}{5.7} 
& \dd{20.0}{8.5} 
& \ddbf{46.46}{10.13} \\

DISCO w/o Policy 
& \dd{65.9}{6.2} 
& \dd{6.7}{2.3} 
& \dd{7.56}{1.59}
& \dd{75.2}{5.0} 
& \dd{33.3}{8.9} 
& \dd{3.02}{0.56} \\

DISCO w/o PW 
& \dd{39.6}{7.2} 
& \dd{0.0}{0.0} 
& \dd{133.14}{31.88}
& \dd{64.8}{7.5} 
& \dd{26.7}{10.8} 
& \dd{22.98}{3.79} \\

\cellcolor{tablecolor}\textbf{DISCO (Ours)} 
& \ddbf{90.9}{6.7} 
& \ddbf{80.0}{9.5} 
& \ddbf{463.16}{79.76}
& \ddbf{94.7}{2.0} 
& \ddbf{66.7}{5.5} 
& \dd{34.12}{5.78} \\

\bottomrule
\end{tabular}}

\vspace{5pt}
\captionof{table}{\small \textbf{Simulation Metrics}. We report average reward and success rate for DISCO against baselines on a held-out test set for shelf packing and object stacking. We also report number of plans found by each method during training. Results are averaged over 32 environment seeds and show a 95\% confidence interval.}
\label{tab:results_main}
\vspace{-15pt}
\end{minipage}
\end{figure}

We compare our method to the following baselines:

\textbf{Beam Search.} We set the beam width to $16$, rank beams greedily by cumulative progress reward rather than a learned value, and distill the policy and value function from the resulting trees.

\textbf{MPPI (Model Predictive Path Integral control).} We replace MCTS with MPPI, which uses the same pretrained 3DFA~\cite{3dfa} policy to sample actions, and learn the value function from its final plans.

\textbf{RL-PD~\cite{rl-pd}.} An off-policy method that augments soft actor-critic with prior data.

\textbf{DISCO w/o Policy.} Instead of sampling actions from the policy, this variant samples heuristic grasp poses for each object and uniformly samples placement poses within the shelf bounds.

\textbf{DISCO w/o PW.} We ablate progressive widening (PW) by sampling a fixed $K{=}16$ children when a vertex is first expanded and adding no new children on subsequent visits.

\textbf{Test-time Planners.} We compare DISCO to test-time planners that use relational state abstractions: SPOT~\cite{saha2025spot}, Points2Plans (P2P)~\cite{huang2024points2plans}, eRDTransformer (eRD-T)~\cite{huang2023erdtransformer}, and Pair-wise RD (P-RD)~\cite{pairwiserd}.

Beam Search, MPPI, and the DISCO ablations replace MCTS only as train-time plan generators. At test time, all methods except the test-time planners, including RL-PD, sample candidate actions from their policy and execute the one ranked highest by their learned value function, without search or a simulator. More details about the baselines can be found in Appendix~\ref{app:baselines}.

\textbf{RQ1: Does DISCO make better long-horizon task progress than search, RL, and sampling-based planners?}
Figure~\ref{fig:results} and Table~\ref{tab:results_main} show that DISCO achieves the highest reward and success on shelf packing and object stacking across train and test tasks, and is the only method to solve $100\%$ of shelf packing and $98\%$ of object stacking train tasks. Beam Search and MPPI improve gradually but plateau below DISCO, suggesting that search without MCTS-style exploration is insufficient, as is sampling-based optimization. RL-PD continues to improve but does not match DISCO, showing that actor-critic updates alone struggle to obtain long-horizon behavior. Since Beam Search greedily follows the progress reward, it also tests the reward alone: it solves only $28\%$ of shelf packing training tasks ($6.7\%$ held-out) versus $100\%$ ($80.0\%$) for DISCO (Appendix~\ref{app:baseline_interpretation}).

\textbf{RQ2: How important are progressive widening and the learned policy prior?}
Figure~\ref{fig:results} and Table~\ref{tab:results_main} show that DISCO's components address bottlenecks in continuous-action tree search. DISCO w/o Policy drops in performance substantially, especially as object count increases, showing that the policy prior is important for focusing search. DISCO w/o PW obtains reasonable training reward and success, but finds fewer successful training plans and degrades sharply in shelf packing as task complexity increases. Progressive widening is thus important for expanding into new action regions and finding broader, multi-modal supervision for distillation (Appendix~\ref{app:pw_analysis}).

\textbf{RQ3: Does DISCO scale better as task complexity increases?}
Figure~\ref{fig:results} shows that DISCO is the only method that solves all tasks across object counts within the simulation budget, while baseline reward decreases gradually and success decreases sharply as task complexity increases.

\begin{figure*}[t]
\centering
\captionsetup{font=small,skip=2pt}

\begin{minipage}[t]{0.42\textwidth}
\centering
\vspace{0pt}

\includegraphics[width=0.82\linewidth]{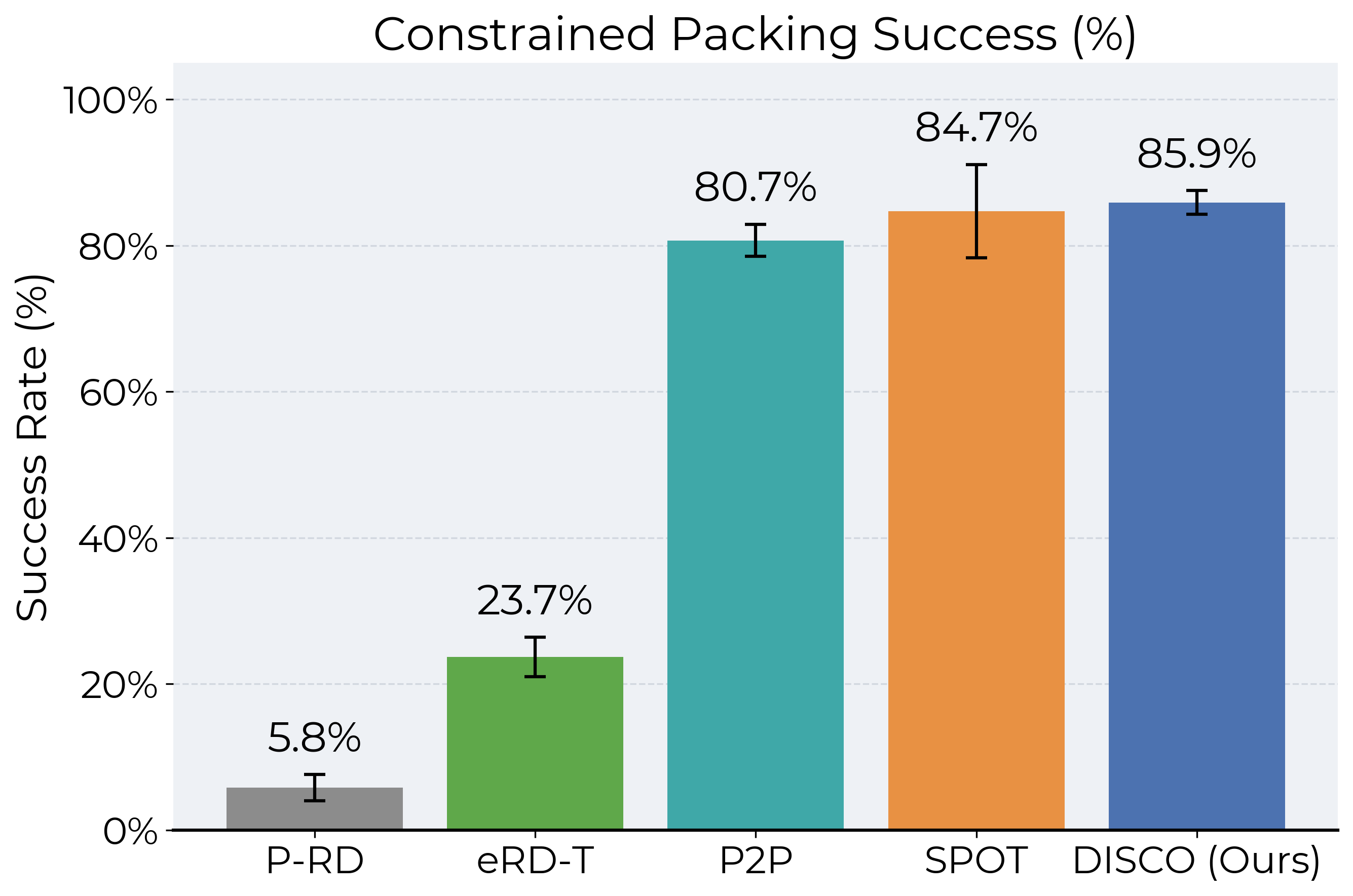}
\captionof{figure}{\textbf{Constrained Packing Metrics.} We rollout DISCO zero-shot against test-time abstraction-based planning methods.}
\label{fig:constrained_packing}

\vspace{8pt}

\resizebox{\linewidth}{!}{%
\begin{tabular}{l|cc|cc}
\toprule
& \multicolumn{2}{c|}{\textbf{Shelf Packing (Real)}}
& \multicolumn{2}{c}{\textbf{Table Bussing (Real)}} \\[3px]
Algorithm
& Reward ($\uparrow$)
& Success ($\uparrow$)
& Reward ($\uparrow$)
& Success ($\uparrow$) \\
\midrule

3DFA 
& \dd{33.3}{19.2} 
& \cc{0/3} 
& \dd{27.8}{5.6}
& \cc{0/3} \\

\cellcolor{tablecolor}\textbf{DISCO (Ours)} 
& \ddbf{88.9}{11.1} 
& \ccbf{2/3} 
& \ddbf{83.3}{9.6}
& \ccbf{1/3} \\

\bottomrule
\end{tabular}}

\vspace{2pt}
\captionof{table}{\textbf{Real World Metrics.} We report average reward and success rate across 3 trials of the shelf packing and table bussing task for our method compared to 3DFA.}
\label{tab:real_world_results}

\end{minipage}
\hfill
\begin{minipage}[t]{0.53\textwidth}
\vspace{0pt}

\includegraphics[width=\linewidth]{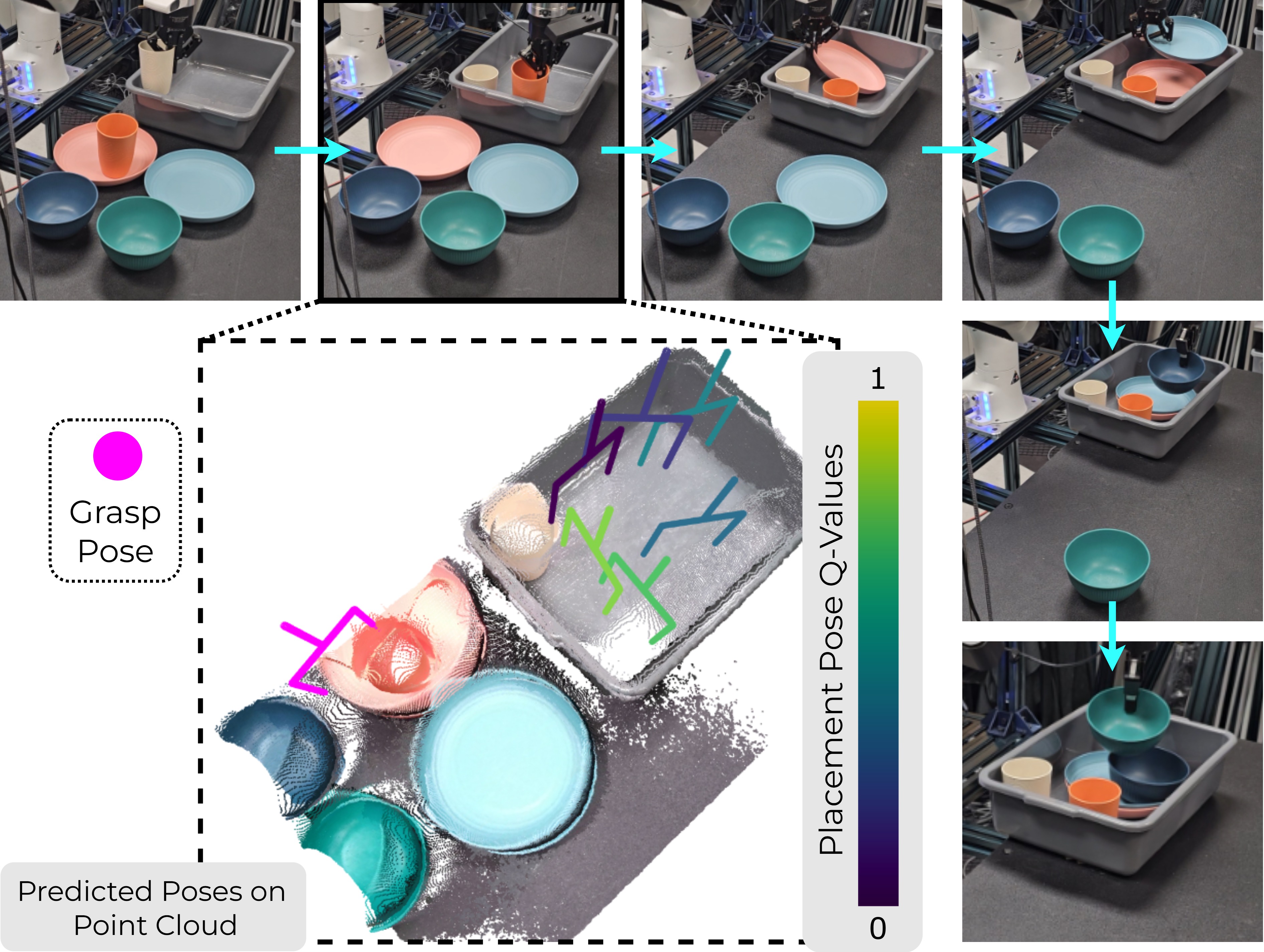}
\vspace{-4pt}
\captionof{figure}{\small \textbf{Real World Rollout} of DISCO on the table bussing task. In the point cloud, we see that DISCO's value function gives high scores for cup placements that leave space for the plates to be placed in the future.}
\label{fig:real_world_qual}

\end{minipage}

\vspace{-10pt}
\end{figure*}

\textbf{RQ4: Does distilling search-generated supervision enable direct rollout during inference without a simulator on held-out tasks?}
On held-out tasks (Table~\ref{tab:results_main}), directly rolling out the distilled policy-value model outperforms all methods on shelf packing and object stacking. On constrained packing (Appendix~\ref{app:constrained_packing_task}), DISCO reaches 85.9\% success under direct rollout, comparable to test-time planners that rely on relational abstractions.
Table~\ref{tab:real_world_results} and Figure~\ref{fig:real_world_qual} show real-world results. On Shelf Packing and Table Bussing, the pretrained 3DFA policy obtains partial reward but fails to complete any rollout, while DISCO improves both reward and success, indicating that its learned long-horizon behavior transfers beyond simulation (Appendix~\ref{app:sim2real}). Since 3DFA is the architecture DISCO distills into, this comparison isolates the effect of the self-improvement loop. In Figure~\ref{fig:real_world_qual}, DISCO's value function favors tray-edge placements over the center, preserving space for larger objects.

\begin{figure}[t]
\centering
\begin{minipage}[t]{0.52\linewidth}
\centering
\includegraphics[width=\linewidth]{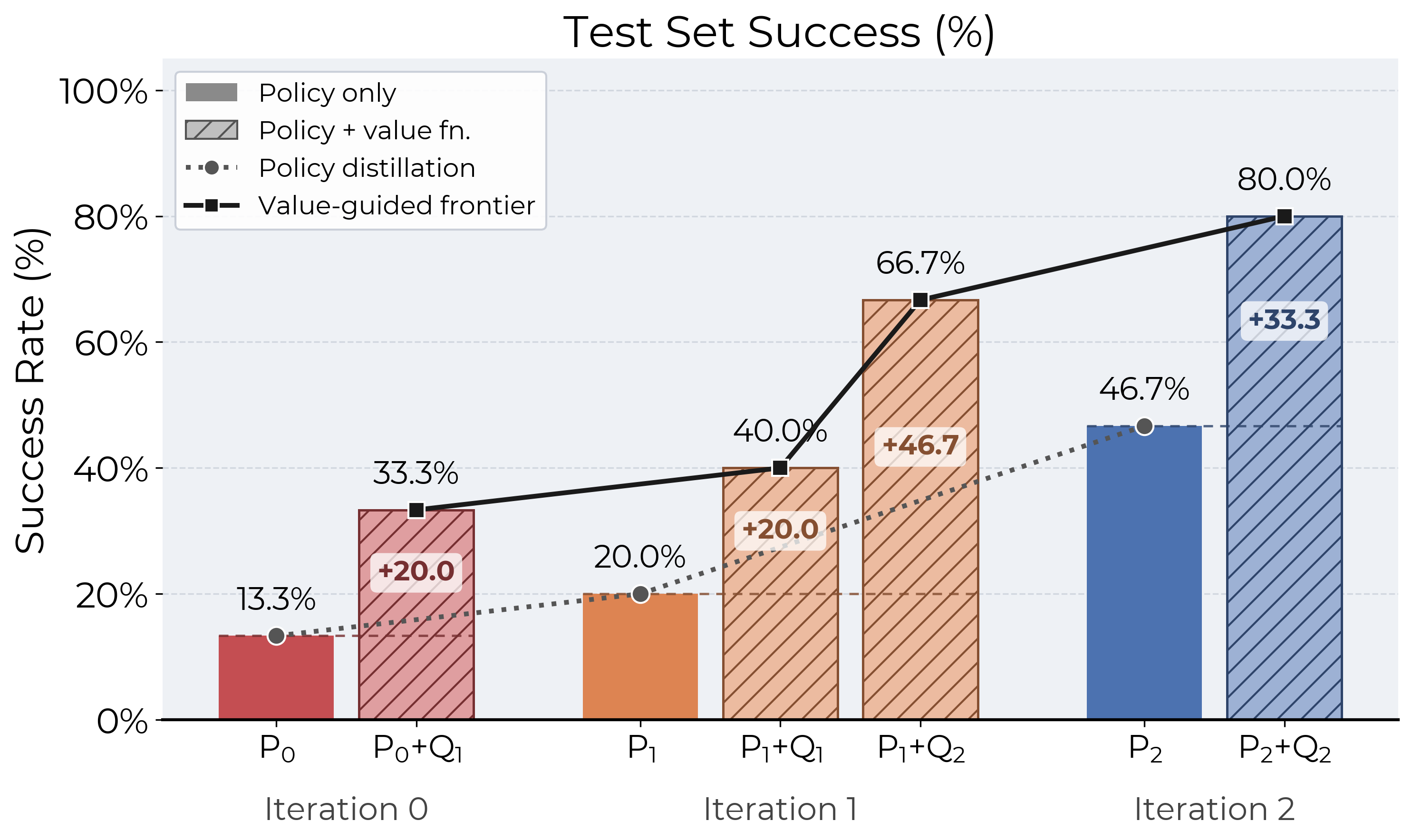}
\captionof{figure}{\small\textbf{Iterative Distillation.} Held-out shelf packing success of the policy $P_k$ after the $k$-th distillation round, executed alone (solid) or with test-time value guidance from $Q_k$ (hatched). Both the policy and the value function improve across rounds.}
\label{fig:distillation_performance}
\end{minipage}
\hfill
\begin{minipage}[t]{0.445\linewidth}
\centering
\includegraphics[width=\linewidth]{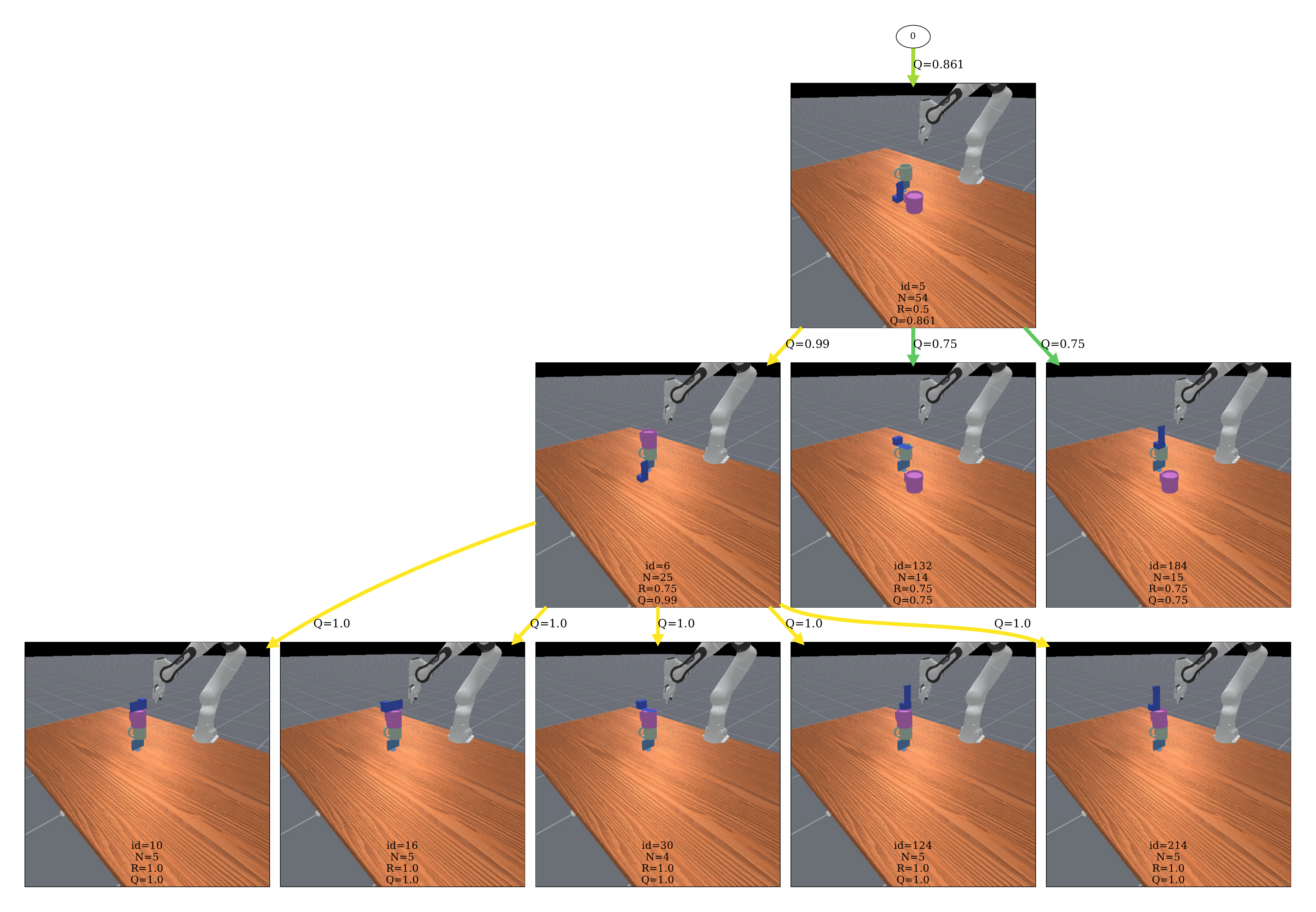}
\captionof{figure}{\small\textbf{Long-Horizon Dependency.} An object stacking search tree in which only placing the pink mug first leads to success, since it supports the blue object placed later. MCTS assigns this action a higher value ($Q=0.99$) than its alternatives ($Q=0.75$).}
\label{fig:funnel_1}
\end{minipage}
\vspace{-10pt}
\end{figure}

\textbf{RQ5: What does iterative distillation learn?}
Figure~\ref{fig:distillation_performance} shows that both networks improve across distillation rounds: the policy alone rises from $13.3\%$ to $46.7\%$ held-out success on shelf packing, and value guidance improves on it at every round, e.g., replacing $Q_1$ with $Q_2$ raises $P_1$ from $40.0\%$ to $66.7\%$ (Appendix~\ref{app:distillation_performance}). In turn, search finds more goal-reaching plans per 25 iterations ($3.36$ to $19.96$ over successive distillation intervals) and reaches greater depths. Figure~\ref{fig:funnel_1} shows the kind of long-horizon dependency that search discovers and single-step demonstrations cannot provide: placing an object only to support later placements (Appendix~\ref{app:qualitative}). 

\section{Conclusion}
We presented DISCO, which bootstraps long-horizon multi-object rearrangement from single-step pick-and-place data by iteratively distilling progressive-widening MCTS over continuous actions into a point-cloud-conditioned policy-value model. Without test-time search, the distilled model outperforms all baselines in simulation and the real world, showing that search is valuable not only as a planner, but also as a source of supervision.
Additionally, iterative distillation provides a self-improving loop. As the policy-value model improves, search discovers more long-horizon plans, which provides richer supervision for the policy, thus improving subsequent searches. Overall, DISCO introduces an algorithm that allows manipulation policies to learn long-horizon reasoning from inexpensive short-horizon data, via a combination of learning and search.


\section{Limitations}
DISCO relies on the initial policy: if it rarely proposes feasible actions, search may not find long-horizon plans to distill. We only evaluate end-effector pick-and-place actions, but since DISCO needs only a proposal policy and a simulator, joint-space or dexterous control is future work. DISCO is also sensitive to simulation fidelity. Although the progress reward is potential-based~\cite{ng1999shaping} and the value targets use only goal indicators, a misaligned progress measure can still bias search. Finally, DISCO does not yet use foundation or vision-language-action models.


\newpage
\acknowledgments{This material is based upon work supported by the Toyota Research Institute, ONR MURI N00014-24-1-2748, and ONR MURI N00014-26-1-2024. This work was also supported by IITP funded by the Korean Government (MSIT) (No.~RS-2024-00457882, National AI Research Lab Project). Additionally, we thank Yishu Li for feedback on the writing, figures and framing of this paper. We also thank Sriram Krishna, Itamar Mishani, Nayesha Gandotra, Pratik Bhowal and Mino Nakura for helpful discussions and feedback on this paper. We are grateful to Haotian Zhan and Vera Xiao for their help with real-world camera calibration. David Held holds concurrent appointments at CMU and as an Amazon Scholar. This paper describes work performed at CMU and is not associated with Amazon.}

\bibliography{references}

\clearpage
\appendix

\setcounter{part}{0}
\renewcommand{\thepart}{}
\part{Appendix}
\parttoc

\section{Baseline Details}
\label{app:baselines}

\subsection{Beam Search}
\label{app:beam_search}

We replace the MCTS planner in DISCO with beam search to test whether MCTS-style exploration and value backups are necessary for long-horizon task progress. Beam search maintains a beam of vertices and expands the highest-scoring vertices at each step. We set the beam width to $16$, matching DISCO's initial branching factor, and run search until a successful leaf is found or the simulation budget is exhausted. Unlike MCTS, beam search has no exploration bonus and does not refine vertex values through repeated backups. The policy and value function are distilled from the beam-search-generated trees after search.
Beam Search is run up to a maximum horizon equal to the number of objects in the task, and terminates early if the goal condition is reached, no feasible successors remain, or the simulation budget is exhausted.
Beam Search uses the same action proposal policy as DISCO, but replaces MCTS selection and progressive widening with fixed-width breadth-limited search. For each beam element, we expand $16$ candidate actions and execute the feasible actions in simulation. Each resulting successor is scored using the cumulative progress reward along the partial plan,
\[
    R(\rho_{0:k})=\sum_{t=0}^{k-1} r(s_t,\mathbf{a}_t,s_{t+1}).
\]
At each search depth, we retain the top beam elements ranked by this cumulative progress reward, with ties broken randomly. Beam Search does not use the learned value function $Q_\phi$ for action selection or ranking.

\subsection{MPPI}
\label{app:mppi}

We replace MCTS with Model Predictive Path Integral (MPPI) control to test whether sampling-based trajectory optimization is sufficient for long-horizon task progress without explicit tree search. MPPI uses the same task setup, action representation, progress reward, and maximum task horizon as DISCO, but removes MCTS selection, progressive widening, and value backups. The value function is not used during MPPI planning. At planning step $k$, MPPI samples a population of $B=32$ candidate action rollouts. We denote the $b$-th rollout by
$$
    \rho^{(b)}_{k:k+H}
    =
    [s_k,\mathbf{a}^{(b)}_k,s^{(b)}_{k+1},\ldots,
    \mathbf{a}^{(b)}_{k+H-1},s^{(b)}_{k+H}],
$$
where $H$ is the planning horizon and each transition satisfies $s^{(b)}_{t+1}=f_{\mathrm{sim}}(s^{(b)}_t,\mathbf{a}^{(b)}_t)$. The return $R(\rho^{(b)}_{k:k+H})$ of a rollout is the cumulative progress reward,
$$
    R(\rho^{(b)}_{k:k+H})
    =
    \sum_{t=k}^{k+H-1}
    r(s^{(b)}_t,\mathbf{a}^{(b)}_t,s^{(b)}_{t+1}).
$$
The planning horizon is equal to the number of objects in the task. Candidate rollouts are weighted using the MPPI exponential return weighting with temperature $\lambda=10.0$,
$$
    w^{(b)}
    =
    \frac{\exp(\lambda R(\rho^{(b)}_{k:k+H}))}
    {\sum_{j=1}^{B}\exp(\lambda R(\rho^{(j)}_{k:k+H}))}.
$$
The policy network is updated from the weighted samples, and the selected action is executed before replanning. We run MPPI until the goal condition is reached, no feasible action can be executed, or the simulation budget is exhausted. The final set of MPPI-generated plans is used as supervision for training the policy-value model under the same distillation and evaluation protocol as DISCO.

\subsection{RL-PD}
\label{app:rl_pd}

We compare against RL-PD~\cite{rl-pd}, an off-policy reinforcement-learning baseline that augments soft actor-critic with prior data. This baseline tests whether long-horizon behavior can be learned through actor-critic updates when the policy is initialized with the same short-horizon supervision used by DISCO. Unlike DISCO, RL-PD does not perform tree search, progressive widening, or MCTS-based distillation. Instead, it learns a policy and critic directly from prior demonstrations and online interaction. RL-PD uses the same observation space, pick-and-place action representation, progress reward, and maximum task horizon as DISCO. We initialize the prior dataset using the same single-step demonstration dataset used to pre-train the DISCO policy. During training, the actor $\pi_\theta(\mathbf{a}_k|\mathbf{o}_k)$ interacts with the simulator using the pick-and-place motion planner. Each executed action produces a transition $(\mathbf{o}_k,\mathbf{a}_k,r_k,\mathbf{o}_{k+1},\mathcal{G}(s_{k+1}))$, where $r_k=r(s_k,\mathbf{a}_k,s_{k+1})$ is the progress reward and $\mathcal{G}(s_{k+1})$ indicates whether the next state satisfies the goal condition. These transitions are stored in a replay buffer together with the prior data. The critic $Q_\phi(\mathbf{o},\mathbf{a})$ is trained to estimate the expected discounted return under the current policy. For a minibatch of transitions, RL-PD uses the soft actor-critic target
$$
    y_k
    =
    r_k
    +
    \gamma \left(1-\mathcal{G}(s_{k+1})\right)
    \left[
        Q_\phi(\mathbf{o}_{k+1},\mathbf{a}_{k+1})
        -
        \alpha \log \pi_\theta(\mathbf{a}_{k+1}|\mathbf{o}_{k+1})
    \right],
$$
where $\mathbf{a}_{k+1}\sim\pi_\theta(\cdot|\mathbf{o}_{k+1})$. The actor is updated to maximize the critic value while maintaining entropy regularization. This provides a reinforcement-learning baseline that uses the same reward signal as DISCO, but obtains long-horizon supervision through temporal-difference learning rather than search-generated plans. We use the hyperparameters shown in Table~\ref{tab:app_rlpd_hyperparameters}. The learned actor-critic is evaluated by direct rollout on held-out tasks without test-time search.

\begin{table}[t]
\centering
\small
\begin{tabular}{l c}
\toprule
Hyperparameter & Value \\
\midrule
Prior data & Single-step demonstrations used by DISCO \\
Discount factor $\gamma$ & $0.99$ \\
Entropy temperature $\alpha$ & $0.1$ \\
Batch size & $64$ \\
Gradient updates per environment iteration & $10$ \\
Critic learning rate & $0.1$ \\
Actor learning rate & $0.01$ \\
Replay buffer capacity & $10^6$ \\
\bottomrule
\end{tabular}
\vspace{4pt}
\caption{\small \textbf{RL-PD Hyperparameters.} We report the actor-critic hyperparameters used for the RL-PD baseline. RL-PD uses the same progress reward, action representation, and task horizon as DISCO, but replaces search-generated supervision with off-policy reinforcement learning from prior and online data.}
\label{tab:app_rlpd_hyperparameters}
\end{table}

\subsection{DISCO without Policy Prior}
\label{app:disco_without_policy}

This ablation removes the learned policy prior from DISCO while keeping the MCTS planner fixed. It tests whether the pretrained policy is necessary for focusing progressive-widening MCTS on useful regions of the continuous action space. The ablation uses the same progress reward, task horizon, simulation budget, progressive-widening rule, UCB selection rule, value-guided rollout, and backup procedure as DISCO. The only change is the action proposal mechanism used when expanding a vertex. In this ablation, we instead construct candidate actions from heuristic grasp poses and uniformly sampled placement poses. 

\textbf{Grasp Pose Sampling}. For each object that has not yet been placed, we retrieve a set of object-frame grasp poses from the stored grasp database from ACRONYM~\cite{acronym} and transform them into the current world frame using the object's pose in the state $s_i$. We then run inverse kinematics (IK) using cuRobo~\cite{curobo} for the transformed grasp poses and keep only IK-feasible grasps. This produces a set of feasible grasp candidates $\{\mathbf{g}^{(m)}_i\}_{m=1}^{M_i}$. where each $\mathbf{g}^{(m)}_i$ is a 7D grasp pose associated with an object index, with a position and a quaternion.

\textbf{Placement Pose Sampling}. For each sampled grasp, we sample placement poses uniformly within the task placement bounds.\\[3px] 
(1) For \underline{Shelf Packing}, the placement position is sampled inside the shelf region after applying the same depth, width, and height tolerances used when executing the place action. The placement orientation is initialized to align the gripper with the shelf and is randomly perturbed around the local axes. This gives a placement set
$\{\mathbf{p}^{(n)}_i\}_{n=1}^{P_i}$
where each $\mathbf{p}^{(n)}_i$ is a 7D placement pose. A candidate action is formed by pairing a heuristic grasp with a uniformly sampled placement,
$\mathbf{a}^{(b)}_i
    =
    (\mathbf{g}^{(m)}_i,\mathbf{p}^{(n)}_i)
$
The resulting action set is executed through the same pick-and-place motion planner used by DISCO. Feasible executions instantiate new child vertices, while infeasible executions are discarded. \\[3px]
(2) For \underline{Object Stacking}, the uninformed sampler uses a geometric stack-action parameterization rather than shelf-bounded placement sampling. A candidate action selects a base object $x_{\mathrm{base}}$, a moved object $x_{\mathrm{move}}$, a face index $f\in\{1,\ldots,6\}$ specifying which oriented bounding-box face of the moved object faces downward, and an offset $(\Delta x,\Delta y)$ on the top face of the base object. Objects already in the current stack are excluded from the moved-object pool. For each base--moved pair and face orientation, $(\Delta x,\Delta y)$ is sampled uniformly within the base object's top-face bounds after subtracting the footprint of the moved object under face orientation $f$. If the moved-object footprint is larger than the base support region, the sampler places the object at the center of the base. The placement height is computed from the base height, the moved object's down-facing dimension, and a small drop gap. The placement orientation composes the base object's orientation with the selected face-down rotation, so the moved object is aligned with the current stack frame. The sampler then checks whether the placement can be reached with a feasible grasp. For each moved object, object-frame grasps are loaded from a cached grasp set and composed with both the current pick pose and the proposed place pose. Candidate grasp--placement pairs are filtered by gripper translation and rotation thresholds and then checked with batched inverse kinematics for both the pick and place poses. The resulting feasible placements are used as the candidate actions for the no-policy-prior ablation in Object Stacking. 

The planner still expands vertices using the progressive-widening condition
$
    |\mathrm{children}(v_i)| < kN(v_i)^\alpha,
$
where $N(v_i)$ is the visit count of vertex $v_i$. Child selection is unchanged from DISCO and uses the UCB score.
We use the same progressive-widening parameters as DISCO, with $k=2.0$ and $\alpha=0.3$, and the same UCB exploration constant $C=1.5$. Search terminates when a state satisfies the goal condition $\mathcal{G}(s)=1$, no feasible successors remain, or the simulation budget is exhausted. This ablation preserves the MCTS search structure and isolates the effect of replacing the learned action prior with heuristic grasp-and-place sampling.

\subsection{Test-Time Planning Baselines}
\label{app:test_time_planners}
We compare DISCO against test-time planning methods on the constrained packing task. These baselines include SPOT, Points2Plans~\cite{huang2024points2plans}, eRDTransformer~\cite{huang2023erdtransformer}, and Pair-wise RD~\cite{pairwiserd}. Unlike DISCO, these methods perform explicit planning at test time using object-centric or relational state representations. This comparison evaluates whether a distilled policy-value model can perform competitively with planners that retain search or relational reasoning during inference.

\textbf{Evaluation protocol.}
The released Points2Plans test set contains five example scenes. We evaluate each scene with $100$ random seeds, producing $500$ total constrained packing trials. We report the success rate over these trials. The test-time planning baselines are evaluated using their planning procedure during inference, while DISCO is evaluated by direct policy-value rollout without simulator-based search.

\textbf{Motion Planning in Points2Plans.}
To make the comparison focus on high-level planning rather than differences in low-level execution, we use a common grasp-and-place execution interface. We modify the Points2Plans action execution to better align with our setup. Instead of using MoveIt inverse kinematics, we compute the Jacobian for the delta pose between the current end-effector pose and the grasp pose predicted by DISCO. We then apply joint velocities in a control loop until the robot reaches the grasp pose. The controller attempts the grasp poses predicted by DISCO. 

\section{Method Details}
\label{app:algorithm_details}

\subsection{Algorithm Details}
\label{app:algorithm_functions}

This section expands the implementation details behind Algorithm~\ref{alg:disco}. We explain algorithmically the mechanism behind the functions \textsc{UCBSelect}, \textsc{Expand}, \textsc{Rollout}, \textsc{Backpropagate} and \textsc{Distill} as they are used in Algorithm~\ref{alg:disco}. The graph notation is as follows: each vertex $\mathbf{v}_i$ stores a simulator state $\mathbf{s}_i$, observation $\mathbf{o}_i$, parent pointer, incoming edge, visit count $N(\mathbf{v}_i)$, and search value $Q(\mathbf{v}_i)$. Each edge stores the action $\mathbf{a}_i$, feasibility status, and the trajectory returned by the motion planner. The set $V_{\mathrm{succ}}$ stores terminal vertices whose states satisfy the goal condition $\mathcal{G}(\mathbf{s}_i)=1$.

\begin{figure*}[t]
\centering
\begin{minipage}[t]{0.46\textwidth}
\vspace{0pt}

\begin{algorithm}[H]
\footnotesize
\DontPrintSemicolon
\caption{\textsc{Expand}}
\label{alg:app_expand}
\KwIn{$G=\langle V,E\rangle$, vertex $\mathbf{v}$, policy $\pi_\theta$, simulator $f_{\mathrm{sim}}$, batch size $B$}
\KwOut{Updated graph $G$}
Construct $\mathbf{o}_{\mathbf{v}}$ from $\mathbf{s}_{\mathbf{v}}$\;
\tcp{batched expansion}
Sample $\{\mathbf{a}^{(b)}\}_{b=1}^{B}\sim \pi_\theta(\mathbf{a}|\mathbf{o}_{\mathbf{v}})$\;
$\{\tau^{(b)}\}_{b=1}^{B} \leftarrow \textsc{MotionPlan}(\mathbf{s}_{\mathbf{v}}, \{\mathbf{a}^{(b)}\}_{b=1}^{B})$\;
$\{\mathbf{s}^{(b)}\}_{b=1}^{B} \leftarrow f_{\mathrm{sim}}(\mathbf{s}_{\mathbf{v}}, \{\tau^{(b)}\}_{b=1}^{B})$\\
\For{$b=1,\ldots,B$}{
    \If{$\tau^{(b)}$ is feasible}{
        Create child $\mathbf{v}^{(b)}$ with state $\mathbf{s}^{(b)}$\;
        Initialize $Q(\mathbf{v}) \leftarrow Q_{\phi}(\mathbf{o}_{\mathbf{v}}, \mathbf{a}^{(b)}) $
        $e^{(b)} \leftarrow \langle \mathbf{v}, \mathbf{a}^{(b)}, \tau^{(b)}, \mathbf{v}^{(b)} \rangle$\;
        $G \leftarrow \langle V\cup\{\mathbf{v}^{(b)}\}, E\cup\{e^{(b)}\}\rangle$\;
    }
}
\Return{$G$}\;
\end{algorithm}

\end{minipage}
\hfill
\begin{minipage}[t]{0.46\textwidth}
\vspace{0pt}

\begin{algorithm}[H]
\footnotesize
\DontPrintSemicolon
\caption{\textsc{Rollout}}
\label{alg:app_rollout}
\KwIn{$G=\langle V,E\rangle$, vertex $\mathbf{v}$, policy $\pi_\theta$, value function $Q_\phi$, reward $r$, goal $\mathcal{G}$, horizon $H$, batch size $B$}
\KwOut{return $R$, terminal vertex $\mathbf{v}_T$, updated graph $G$}
$\mathbf{v}_{\mathrm{cur}} \leftarrow \mathbf{v}$\;
$R \leftarrow 0$\;
\For{$h=1,\ldots,H$}{
    \If{$\mathcal{G}(\mathbf{s}_{\mathbf{v}_{\mathrm{cur}}})=1$}{
        \textbf{break}\;
    }
    $G \leftarrow \textsc{Expand}(G,\mathbf{v}_{\mathrm{cur}},\pi_\theta,f_{\mathrm{sim}},B)$\;
    Construct $\mathbf{o}_{\mathrm{cur}}$ from $\mathbf{s}_{\mathbf{v}_{\mathrm{cur}}}$, get actions $\{\mathbf{a}^{(b)}\}_{b=1}^{B}$ from $\mathrm{children}(\mathbf{v}_{\mathrm{cur}})$\;
    $\mathbf{v}_{c}^{*}
    \leftarrow
    \arg\max
    \textsc{UCB}(\mathbf{v}, C)$\;
    $r_i \leftarrow r(\mathbf{s}_{\mathbf{v}_{\mathrm{cur}}},\mathbf{a}^{*},\mathbf{s}_{\mathbf{v}_{c}^{*}})$\;
    $R \leftarrow R+r_i$\;
    $\mathbf{v}_{\mathrm{cur}} \leftarrow \mathbf{v}_{c}^{*}$\;
}
$\mathbf{v}_T \leftarrow \mathbf{v}_{\mathrm{cur}}$\;
\Return{$R,\mathbf{v}_T,G$}\;
\end{algorithm}

\end{minipage}
\vspace{-6pt}
\end{figure*}

\begin{figure*}[b]
\centering
\begin{minipage}[t]{0.46\textwidth}
\vspace{0pt}

\begin{algorithm}[H]
\footnotesize
\DontPrintSemicolon
\caption{\textsc{UCBSelect}}
\label{alg:app_ucb_select}
\KwIn{Search graph $G=\langle V,E\rangle$, root vertex $\mathbf{v}_0$, exploration constant $C$}
\KwOut{Leaf vertex $\mathbf{v}$}
$\mathbf{v} \leftarrow \mathbf{v}_0$\;
\While{$\mathrm{children}(\mathbf{v}) \neq \emptyset$}{
    $\mathbf{v}_{c}^{*}\leftarrow
    \arg\max
    \textsc{UCB}(\mathbf{v}, C)
    $\;
    $\mathbf{v} \leftarrow \mathbf{v}_{c}^{*}$\;
}
\Return{$\mathbf{v}$}\;
\end{algorithm}

\end{minipage}
\hfill
\begin{minipage}[t]{0.46\textwidth}
\vspace{0pt}

\begin{algorithm}[H]
\footnotesize
\DontPrintSemicolon
\caption{\textsc{Backpropagate}}
\label{alg:app_backprop}
\KwIn{$G$, terminal vertex $\mathbf{v}_T$, return $R$}
\KwOut{updated $G$}
$\mathbf{v} \leftarrow \mathbf{v}_T$\;
\While{$\mathbf{v}\neq \emptyset$}{
    $N(\mathbf{v}) \leftarrow N(\mathbf{v}) + 1$\;
    $Q(\mathbf{v}) \leftarrow
    Q(\mathbf{v}) +
    \frac{R-Q(\mathbf{v})}{N(\mathbf{v})}$\;
    \If{$\mathcal{G}(\mathbf{s}_{\mathbf{v}})=1$}{
        $V_{\mathrm{succ}} \leftarrow
        V_{\mathrm{succ}}\cup\{\mathbf{v}\}$\;
    }
    $\mathbf{v} \leftarrow \mathrm{parent}(\mathbf{v})$\;
}
\Return{$G$}\;
\end{algorithm}

\end{minipage}
\vspace{-6pt}
\end{figure*}
\textbf{\textsc{UCBSelect}.} This function traverses the current tree (with expanded nodes) from the root until it reaches a vertex, which will later be selected for expansion or rollout. The UCB selection rule uses stored visit counts and vertex values in the search graph, following standard MCTS. The procedure is illustrated in Algorithm~\ref{alg:app_ucb_select}\\[3px]
\textbf{\textsc{Expand}}.
This function adds new children to the selected vertex when the progressive-widening condition is satisfied. Given a vertex $\mathbf{v}$, we construct its point-cloud observation $\mathbf{o}_{\mathbf{v}}$ and sample candidate actions from $\pi_\theta(\mathbf{a}|\mathbf{o}_{\mathbf{v}})$. For each sampled action $\mathbf{a}^{(b)}$, the motion planner first produces a trajectory $\tau^{(b)}$ from the current robot state. Feasible trajectories are executed in simulation to obtain the next state $\mathbf{s}^{(b)} = f_{\mathrm{sim}}(\mathbf{s}_{\mathbf{v}}, \tau^{(b)})$. Each feasible transition instantiates a new child vertex $\mathbf{v}^{(b)}$ with state $\mathbf{s}^{(b)}$ and an incoming edge storing $\mathbf{a}^{(b)}$ and $\tau^{(b)}$ as $e^{(b)} = \langle \mathbf{v}, \mathbf{a}^{(b)}, \tau^{(b)}, \mathbf{v}^{(b)} \rangle$. Infeasible motion plans are discarded. This keeps policy supervision restricted to executable state-action transitions. The new vertices and edges are added to the graph. These new vertices are scored in batch by $Q_{\phi}$, thus initializing the child's value.
$
G \leftarrow \langle V \cup \{\mathbf{v}^{(b)}\}, E \cup \{e^{(b)}\} \rangle.
$\\[3px]
\textbf{\textsc{Rollout}}.
Algorithm~\ref{alg:app_rollout} rolls out from the a newly expanded vertex into a short simulated trajectory. At each rollout step, \textsc{Expand} samples a batch of actions from the policy, filters them through motion planning and simulation, and adds the feasible children to the graph. The rollout then selects the best child using the learned value function and UCB exploration term. The selected edge reward is added to the cumulative return $R$, and rollout continues from the selected child until the goal is reached, the horizon is exhausted, or no feasible child is available. This adds feasible rollout transitions to the search graph, providing additional state-action supervision for distillation. At test time, DISCO uses the same policy-value scoring interface without graph construction, UCB exploration, or $f_{\mathrm{sim}}$. The selected action is executed on the robot, the next observation is received, and the policy-value model is queried again.\\[3px]
\textbf{\textsc{Backpropagate}.}
Algorithm~\ref{alg:app_backprop} propagates the observed rollout return through the parent chain. Each visited vertex updates its visit count and running-average value estimate. Successful terminal vertices are added to $V_{\mathrm{succ}}$, allowing their parent chains to be recovered as plans for policy-value distillation.\\[3px]
\textbf{\textsc{Distill}.} The procedure for this is illustrated in the iterative distillation section in the main paper.

\subsection{MCTS Hyperparameters}
\label{app:mcts_hyperparameters}

Hyperparameters used are shown in Table~\ref{tab:app_mcts_hyperparameters}. The simulation budget $\beta$ caps the number of simulator calls used across the train set, which we set to 270,000 for shelf packing and 27,000 for object stacking. There are $B=32$ candidates per expansion which gives enough headroom for the hardest tasks. Progressive widening ($k$, $\alpha$)
controls how many children a node may have as a function of its visit count,
which is required in our continuous action space to keep branching bounded
and concentrate simulation on promising subtrees. The exploration constant
$C$ trades off exploitation and exploration in the UCB rule. Since returns
$R=\kappa(s_T)-\kappa(s_0)$ lie in $[0,1]$, a value near $\sqrt{2}$ suffices. The expansion
batch size $B$ is the per-expansion cap on policy-proposed (grasp, place)
candidates. The 32 parallel GPU environments are filled with candidates from
concurrent descents, so one simulation batch can contain state-action pairs from different vertices. The task horizon $K$ equals the number of
objects, since each tree level corresponds to one pick-and-place. Trees are
distilled into the policy and $\hat{Q}$ after each complete planning sweep
over the training set ($N_{\mathrm{distill}}$), for two rounds total, after
which planning runs purely from the distilled networks. Each round contains a distillation of the policy, and a distillation of the value function. Training uses 50 tasks, and 15 held-out tasks measure generalization.

\begin{table}[t]
\centering
\small
\begin{tabular}{l c}
\toprule
Hyperparameter & Value \\
\midrule
Simulation budget $\beta$ & 270,000 (shelf packing), 27,000 (object stacking) \\
Progressive-widening coefficient $k$ & 2.0 \\
Progressive-widening exponent $\alpha$ & 0.3 \\
Exploration constant $C$ & 1.5 \\
Expansion batch size $B$ & 32 \\
Maximum task horizon $K$ & number of objects (3--7 per task) \\
Number of distillation rounds & 2 \\
Number of training seeds/tasks & 50 \\
Number of held-out seeds/tasks & 15 \\
\bottomrule
\end{tabular}
\vspace{5pt}
\caption{\small MCTS and distillation hyperparameters.}
\label{tab:app_mcts_hyperparameters}
\end{table}

\subsection{Policy Architecture}
\label{app:policy_arch}

\begin{figure}[t]
    \centering
    \includegraphics[width=\linewidth]{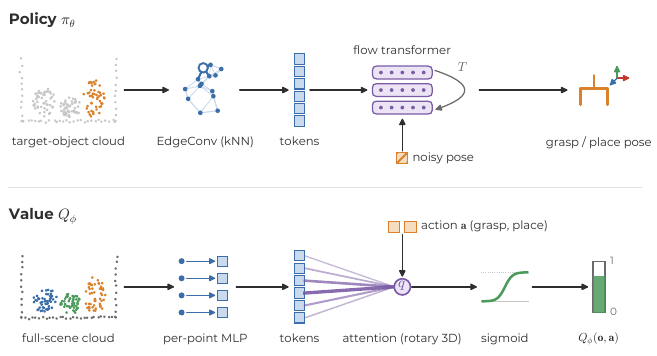}
    \caption{\small\textbf{Policy and Value Function Architectures.} \textbf{Top}: the policy $\pi_\theta$ encodes the target-object point cloud with EdgeConv layers and denoises a pose token with a flow-matching transformer to predict a grasp or place pose. \textbf{Bottom}: the value function $Q_\phi$ embeds the full-scene point cloud with a per-point MLP, and an action-conditioned query attends to the scene tokens to predict the success probability $Q_\phi(\mathbf{o},\mathbf{a})\in[0,1]$.}
    \label{fig:app_policy_value_arch}
\end{figure}

We use a modified 3D FlowMatch Actor (3DFA)~\cite{3dfa} architecture for the policy $\pi_{\theta}$ (Figure~\ref{fig:app_policy_value_arch}, top). In contrast to the original 3DFA model, which uses a CLIP-based visual encoder, we replace the visual backbone with a PointNet++-style point-cloud encoder. The policy operates on an xyz-only point cloud of the scene, with a mask assigned to the object to be grasped and placed. We assume access to this segmentation mask of the target object. The policy then predicts continuous SE(3) end-effector poses. It is implemented as two separately trained rectified-flow actors with the same architecture: a grasp model and a grasp-conditioned place model. The grasp model predicts a single grasp pose, while the place model predicts a single placement pose conditioned on the selected grasp.

The point-cloud encoder maps a 4096-point cloud to 120-dimensional per-point scene tokens using two EdgeConv layers over a shared $k$-nearest-neighbor graph ($k=16$) built in xyz space. The input has 3 channels (xyz) for shelf packing and 4 channels (xyz and a binary target-object mask) for object stacking. These scene tokens are processed by a transformer denoising head with rotary 3D positional encodings. The policy uses timestep conditioning through adaptive layer normalization, and predicts a velocity field for rectified-flow denoising. At inference time, actions are sampled by initializing from Gaussian noise and integrating the predicted velocity field with Euler steps. The output of each policy call is one pose token $[x,y,z,q_w,q_x,q_y,q_z,g]$, where $g$ denotes the gripper state. The grasp model fixes $g=1$, and the place model fixes $g=0$.

The policy is applied in a cascaded manner. For each object that is not already packed, we construct a target-object-only point cloud while keeping the shelf and already packed objects as placement context, thus we run the policy for every object that we have segmented in the scene. The grasp model samples candidate grasps for the target object. Each predicted grasp is passed as conditioning to the place model, which samples placement poses. Each resulting grasp-place pair forms one composite action $\mathbf{a}$ used by MCTS.

\subsection{Value Function Architecture}
\label{app:value_arch}

The value function $Q_{\phi}(\mathbf{o},\mathbf{a})$ scores a point-cloud observation and a grasp-place action with a scalar value in $[0,1]$ (Figure~\ref{fig:app_policy_value_arch}, bottom). Unlike the policy, the value function sees the whole point cloud of the scene (and not just the target object) so that it can reason about long horizon success probability. The action is represented as two pose tokens, one for grasping and one for placing. Each token is embedded with an MLP and a learned type embedding, and the two action tokens are contextualized with self-attention. The scene is encoded from xyz point tokens: each point is embedded independently by a two-layer per-point MLP (without neighborhood aggregation) into a 120-dimensional token, followed by attention layers (4 shared layers, 8 heads) with rotary 3D positional encodings. A learned value query attends to the scene using the action embedding as conditioning. The final value head predicts the probability that search below the resulting state-action pair reaches a successful terminal state. The training objective is shown in the iterative distillation section in the main paper. The value function uses a sigmoid activation to keep predictions between 0 and 1, representing the probability of success.

\begin{figure}[t]
    \centering
    \includegraphics[width=0.9\linewidth]{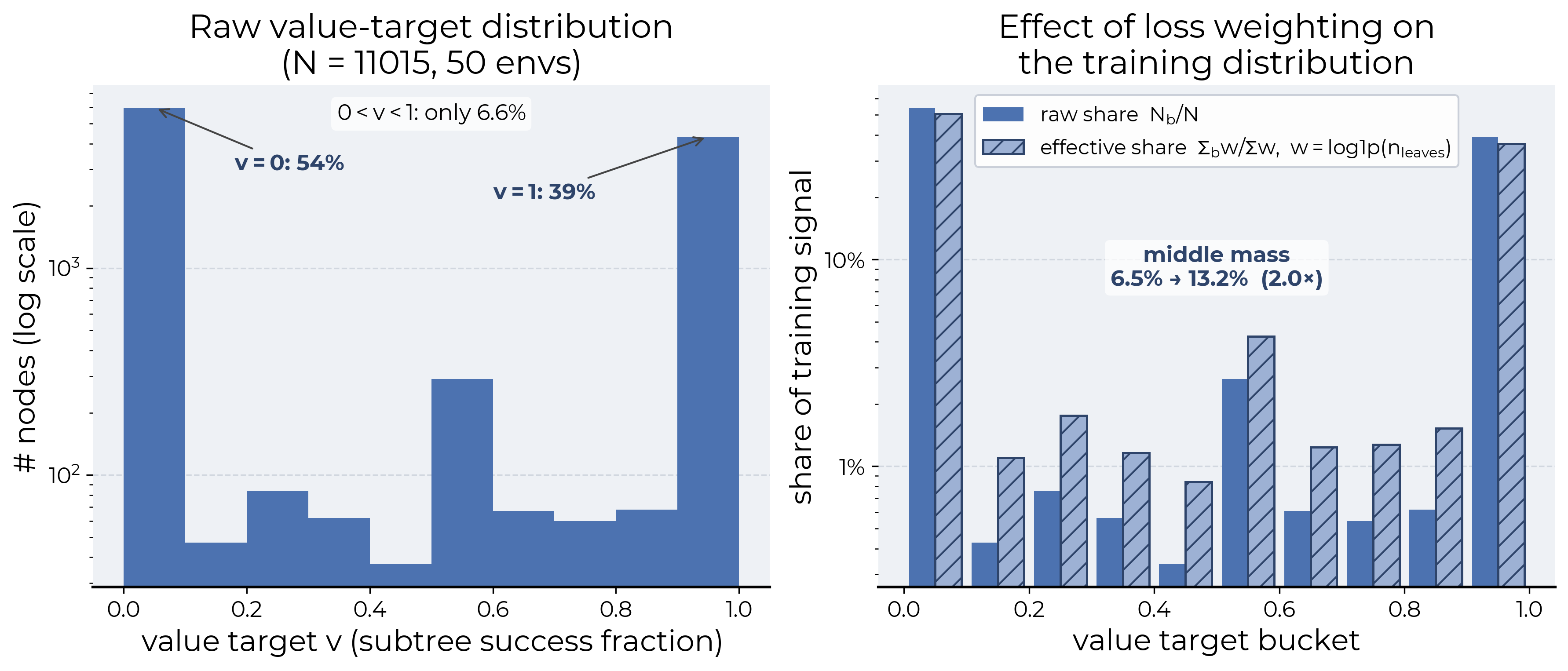}
    \caption{\small\textbf{Value Function Targets.} We visualize the distribution of value-function targets and the effect of evidence weighting. \textbf{Left}: histogram of targets $v=n_{\mathrm{succ}}/n_{\mathrm{leaves}}$ over 11,015 eligible nodes from 50 training trees, showing that the distribution is dominated by saturated targets at 0 and 1. \textbf{Right}: raw target-bucket mass compared to the effective training mass under the weighted BCE loss with $w=\log(1+n_{\mathrm{leaves}})$, which increases the contribution of intermediate targets backed by more rollout evidence.}
    \label{fig:value_target_hist}
\end{figure}

We train the value function using targets extracted from expanded MCTS trees. For each visited non-root vertex, the target is the fraction of successful leaf vertices among its visited descendants, where a leaf is a visited vertex with no visited feasible children (e.g., a vertex at the task horizon or a dead end). The value loss is a weighted binary cross-entropy, with higher weight assigned to targets computed over more leaves. During search, candidate actions are scored in batch by $Q_{\phi}$ at the parent observation (as shown in line 8 of Algorithm~\ref{alg:app_expand}). This predicted value initializes the child value before MCTS backups further refine it through simulated returns.

We weight each per-sample BCE loss $\ell_i$ by $w_i=\log(1+|\mathrm{leaves}(\mathbf{v}_i)|)$ and normalize within the minibatch, giving $\mathcal{L}_V=\sum_i w_i \ell_i / \sum_i w_i$, where $|\mathrm{leaves}(\mathbf{v}_i)|$ is the number of visited leaves in the subtree of vertex $\mathbf{v}_i$. This acts as evidence weighting: targets backed by more rollout outcomes receive larger influence, while the logarithm prevents large subtrees from dominating the loss. Figure~\ref{fig:value_target_hist} shows why this weighting is useful. The left panel shows that the raw target distribution is highly saturated, with most targets exactly 0 or 1. The right panel shows that weighting increases the effective training signal from intermediate targets, which are backed by more rollout evidence and contain the ranking information needed by the value function.

\subsection{Sim-to-Real Augmentation}
\label{app:sim2real}

DISCO uses observations that can be computed both in simulation and on the real robot. Both the policy and value function consume 4096-point xyz clouds in the robot base frame. The policy uses target-object-only clouds for action generation, while the value function scores the resulting action using the corresponding point-cloud observation. We do not use simulator state, proprioception or RGB features as network inputs, which makes sim2real transfer easier. We assume access to segmented point clouds of the object to be picked and placed in both the real world and simulation.

For sim2real transfer, we apply point-cloud augmentation during policy and value training. For each training sample, we add per-point Gaussian noise to xyz coordinates, with the noise scale sampled uniformly up to $0.01$m. We also remove local neighborhoods from the cloud and replace the removed points with uniformly sampled points inside the cloud bounding box. This models missing depth points, segmentation artifacts, and partial occlusions.

We further apply geometric augmentation with separate transforms for object and shelf points. Object points are rotated about the vertical axis, matching the fact that objects rest on a table. Shelf points are transformed with full 3D rotations, since the real shelf pose may differ from simulation. The corresponding grasp and place actions are transformed consistently with the object and shelf transforms, respectively. After transforming the cloud into the robot frame, we canonicalize each sample by subtracting the point-cloud centroid and scaling by the maximum radius. At deployment, the same normalization is applied to the real point cloud and inverted on the predicted actions before execution. This makes the policy-value model less sensitive to absolute camera pose and scene placement.

\subsection{Pre-training Dataset for Initial Policy}
\label{app:pretrain_dataset}
\begin{table}[t]
\centering
\small
\begin{tabular}{llcc}
\toprule
Model & Metric & Training Steps & Final \\
\midrule
Grasp policy & $\mathcal{L}_{\mathrm{pos}}$ ($\downarrow$) & 82.8k & 0.0457 \\
Grasp policy & $\mathcal{L}_{\mathrm{rot}}$ ($\downarrow$) & 82.8k & 0.714 \\
Place policy & $\mathcal{L}_{\mathrm{pos}}$ ($\downarrow$) & 58.9k & 0.0530 \\
Place policy & $\mathcal{L}_{\mathrm{rot}}$ ($\downarrow$) & 58.9k & 0.384 \\
Q function & $\mathcal{L}_{\mathrm{MSE}}^{Q}$ ($\downarrow$) & 30k & 0.00681 \\
Q function & $\mathcal{L}_{\mathrm{MAE}}^{Q}$ ($\downarrow$) & 30k & 0.0160 \\
\bottomrule
\end{tabular}
\vspace{5pt}
\caption{\small\textbf{Policy and Value Training Metrics.} We report final training losses for the grasp policy, place policy, and Q function. The grasp policy is trained for 82.8k steps and the place policy is trained for 58.9k steps using the flow-matching policy objective, where $\mathcal{L}_{\mathrm{pos}}$ denotes translation error and $\mathcal{L}_{\mathrm{rot}}$ denotes rotation error. The Q function is trained for 30k steps, where $\mathcal{L}_{\mathrm{MSE}}^{Q}$ and $\mathcal{L}_{\mathrm{MAE}}^{Q}$ denote mean squared error and mean absolute error on subtree-success targets.}
\label{tab:app_policy_value_train_metrics}
\end{table}
We initialize the policy using single-step grasp-and-place data generated with task-specific geometric samplers. These are the same samplers that we use for the DISCO without Policy Prior baseline in Section~\ref{app:disco_without_policy}. For each sampled simulator state, we first identify the objects that can still contribute to task progress. In Shelf Packing, objects that are already placed in the shelf are removed from the candidate set, and the sampler generates actions only for the remaining objects. Grasps are obtained by transforming stored object-frame grasps into the current world frame and filtering them with inverse kinematics. For each feasible grasp, placement poses are sampled inside the shelf bounds using the same placement tolerances used when executing the place action. In Object Stacking, the sampler removes objects that are already in the current stack from the moved-object set. It then samples stack actions by choosing a base object, a moved object, a downward-facing object face, and an $(\Delta x,\Delta y)$ offset on the top face of the base object. The placement height is computed from the base height and moved-object extent, and the placement orientation aligns the moved object with the selected face-down orientation. Candidate stack placements are paired with cached object grasps and filtered by pick-and-place inverse kinematics. For both tasks, each feasible action is executed in simulation, and successful transitions provide observation-action pairs $(\mathbf{o}_k,\mathbf{a}_k)$ for behavior cloning. Most importantly, this dataset contains local single-step manipulation supervision, but does not contain demonstrations of complete long-horizon task solutions.

\subsection{Policy-value training metrics.}

Table~\ref{tab:app_policy_value_train_metrics} reports training diagnostics for the grasp policy, place policy, and value function. The value function reaches low training error on the normalized subtree-success targets, with final MSE $0.00681$ and MAE $0.0160$, suggesting that it provides useful scores for UCB-based child ranking. The grasp and place policy are executed directly and have a final position error of approximately $4$cm and $5$cm respectively, indicating that there can be multi-modality in the action distribution for grasping and placement. DISCO takes advantage of this multi-modality by sampling a batch of 32 candidates per expansion and filtering infeasible actions through IK and execution, which allows search to produce multiple useful children even when individual samples are noisy.

\section{Task Setup}
\label{app:task_setup}

\subsection{Shelf Packing.}
We implement the task in ManiSkill3~\cite{maniskill}.
In Shelf Packing, the robot must move all objects from the table into a shelf. Each environment contains a procedurally generated shelf and a set of graspable objects sampled from ACRONYM meshes with cached grasp sets. Objects are initialized on the table at random collision-free poses with random yaw rotations. The shelf is modeled as a five-wall box with randomized width, depth, height, floor distance, wall thickness, and small pose perturbations. This produces a distribution of shelf geometries and object layouts for training and evaluation.
\begin{figure}[t]
    \centering
    \includegraphics[width=\linewidth]{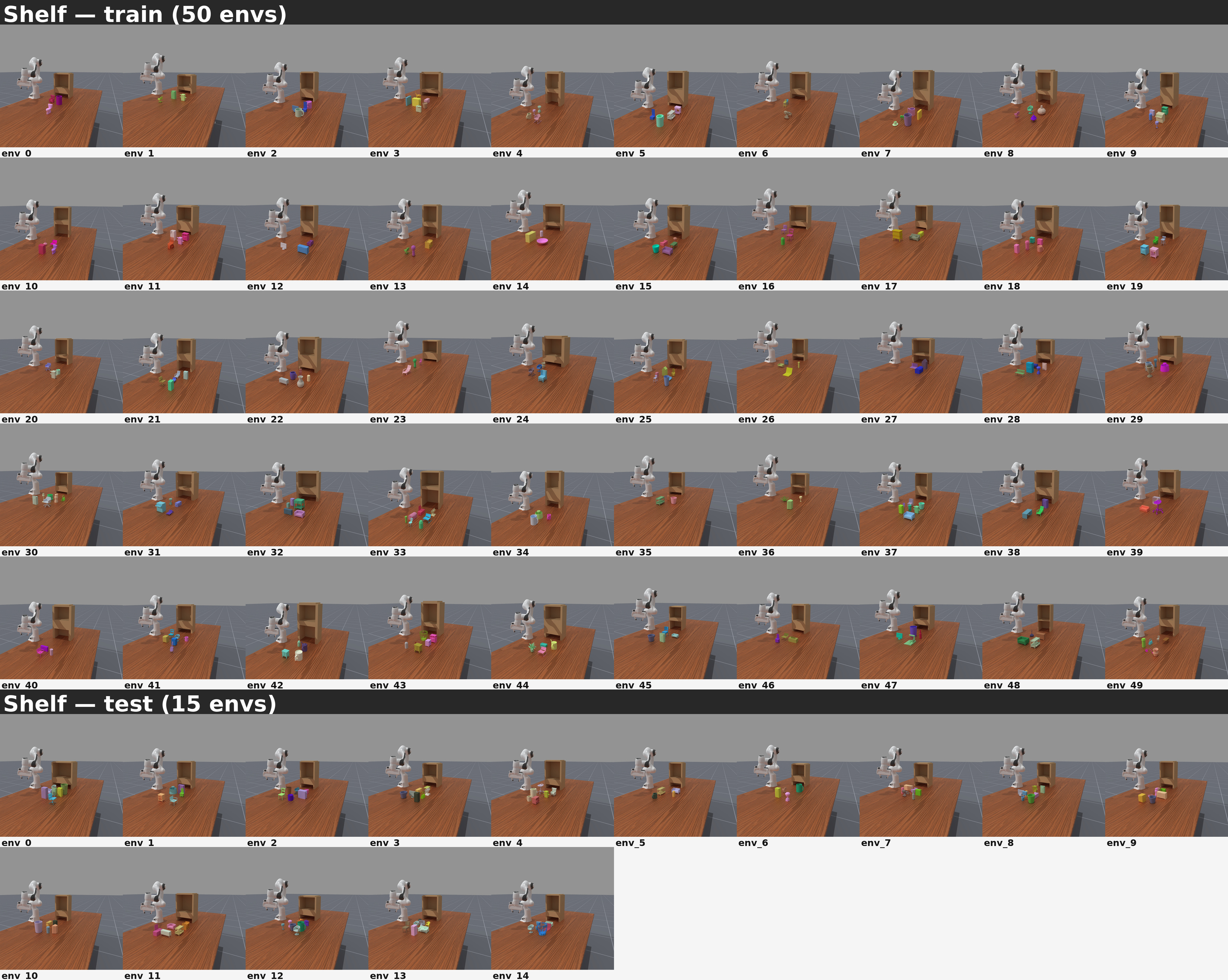}
    \caption{\small\textbf{Shelf Packing Environments}. Visualization of all 50 train environments and 15 test environments for the shelf packing task setup.}
    \label{fig:shelf_packing_envs}
\end{figure}

The task progress $\kappa(s)$ is the packed volume fraction, computed as the total volume of objects whose centers lie inside the shelf cuboid divided by the total object volume. An object is considered inside the shelf when its center is within the shelf bounds in shelf coordinates. The episode succeeds when all objects are inside the shelf. During MCTS, the edge reward $r(s,\mathbf{a},s')=\kappa(s')-\kappa(s)$ is the change in packed fraction. This gives dense task progress while preserving the terminal objective of packing every object.

\subsection{Object Stacking.}
We implement the task in ManiSkill3~\cite{maniskill}. Object Stacking uses the same object layouts as Shelf Packing, but removes the shelf and changes the task objective. Each environment is derived from the corresponding shelf-packing environment by deleting the shelf geometry while keeping the object poses fixed. The action sampler chooses a base object, a moved object, one of six face-down object orientations, and an $(x,y)$ offset on the top face of the base object. The moved object is then placed at the sampled relative pose and the simulator is stepped until the scene stabilizes.
\begin{figure}[t]
    \centering
    \includegraphics[width=\linewidth]{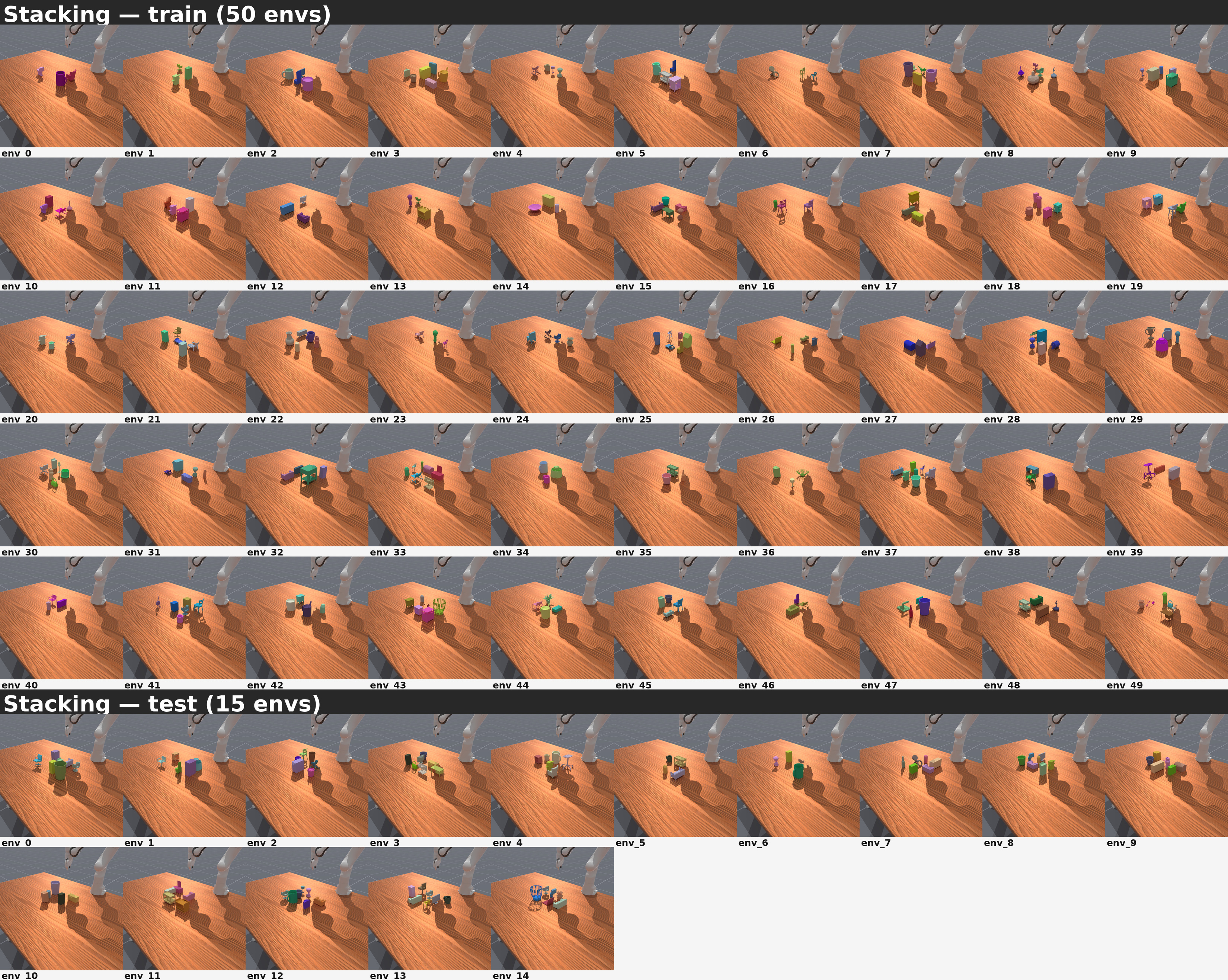}
    \caption{\small\textbf{Object Stacking Environments}. Visualization of all 50 train environments and 15 test environments for the object stacking task setup.}
    \label{fig:object_stacking_envs}
\end{figure}

A moved object is considered stacked on a base object when the two objects are in contact, the top object is horizontally close to the base object, and the top object is vertically above the base. A stacking action is marked successful only if the moved object remains stacked on the selected base and the base object does not drift beyond the allowed translation and rotation tolerances. The task progress $\kappa(s)$ is the normalized tower height, given by the length of the stacked chain divided by the number of objects. The episode succeeds when the resulting tower reaches the required task height.

\subsection{Constrained Packing}
\label{app:constrained_packing_task}

This environment is used to compare direct DISCO rollout against test-time abstraction-based planning methods. In constrained packing, the goal is to arrange all objects on top of the shelf surface without stacking objects on top of each other. We use a success condition that is similar to the Points2Plans evaluation protocol, but adapted to our goal function. Points2Plans defines success using goal predicates and feasibility-related predicates, including predicates that indicate whether an object's placement location is blocked by an object that has already been placed. Since DISCO does not use relational abstractions or symbolic predicates, we do not reproduce this metric exactly. Instead, we define success using the final object arrangement: all objects must lie on the shelf surface, no object should be stacked on another object, and previously placed objects should not be disturbed. This allows objects to be placed behind other objects when the final arrangement remains valid.

\subsection{Real-World Shelf Packing and Table Bussing}
\label{app:real_world_tasks}
Figure~\ref{fig:task_setup} shows the real world environment for shelf packing and table bussing. We evaluate using a Franka Panda robot with a Robotiq gripper. In real-world Shelf Packing, the robot moves objects from the table into the shelf. In Table Bussing, the robot places dinnerware upright in a tray while preserving space for remaining objects. We use the same success conditions and reward conditions for real world shelf packing, as simulated shelf packing. For table bussing, we define task progress as the ratio of objects that are in the tray, and the success condition as all objects being in the tray. An object is considered to be in the tray if its centroid is within the tray's bounding box and it is placed without any tilt, that is it has a roll and pitch angle less than 30 degrees in the global frame.

\section{Additional Quantitative Analysis}
\label{app:additional_analysis}

\subsection{How do distillation iterations improve performance?}
\label{app:distillation_performance}

Figure~\ref{fig:distillation_performance} in the main paper shows held-out test success across distillation iterations $N_{distill}$, separating improvements from the policy and the value function. Here, $P_k$ denotes the policy after the $k$-th distillation iteration, with $P_0$ the base policy, and $Q_k$ denotes the value function trained at iteration $k$. Solid bars show the policy executed alone, while hatched bars show the same policy with test-time value guidance. The dotted line traces the policy-only backbone $P_0 \rightarrow P_1 \rightarrow P_2$, showing that each round of policy distillation improves standalone performance from $13.3\%$ to $20.0\%$ to $46.7\%$. The solid line traces the value-guided frontier, which remains above the policy-only backbone at every stage, indicating that value guidance consistently improves execution. Within each group, the dashed horizontal line marks the corresponding policy-only level, and the in-bar ``+$x.x$'' annotation gives the absolute gain from value guidance at fixed policy. The two guided bars for $P_1$ isolate improvement in the value function itself: keeping the policy fixed, replacing $Q_1$ with $Q_2$ increases success from $40.0\%$ to $66.7\%$. Finally, $P_2$ alone exceeds the previous guided system $P_1+Q_1$, while $P_2+Q_2$ reaches the strongest performance overall. This indicates that iterative distillation improves both the standalone policy and the value-guided rollout behavior.

\subsection{Distillation Ablations}
\label{app:distill_ablations}

\begin{table}[t]
\centering
\begin{minipage}[t]{0.48\linewidth}
\centering
\resizebox{\linewidth}{!}{%
\begin{tabular}{l|cc}
\toprule
Policy trained on & Success (\%) ($\uparrow$) & Dataset size \\
\midrule
Only successful plans & \dd{32.3}{24.9} & 1,868 \\
\cellcolor{tablecolor}\textbf{All feasible edges (Ours)} & \ddbf{52.8}{38.9} & 8,655 \\
\bottomrule
\end{tabular}}
\end{minipage}
\hfill
\begin{minipage}[t]{0.48\linewidth}
\centering
\resizebox{\linewidth}{!}{%
\begin{tabular}{l|cc}
\toprule
Sub-task data & \#Demos & Reward ($\uparrow$) \\
\midrule
25\% of $\mathcal{D}$ & 444 & \dd{28.4}{7.5} \\
50\% of $\mathcal{D}$ & 888 & \dd{32.5}{6.8} \\
75\% of $\mathcal{D}$ & 1,332 & \dd{41.0}{8.0} \\
\cellcolor{tablecolor}\textbf{100\% of $\mathcal{D}$ (Ours)} & 1,776 & \ddbf{41.9}{8.2} \\
\bottomrule
\end{tabular}}
\end{minipage}
\vspace{2pt}
\caption{\small \textbf{Distillation Ablations.} \textbf{Left}: training the policy on all feasible edges gives higher success in subsequent search than training on only successful plans. \textbf{Right}: pre-training on fractions of the sub-task dataset $\mathcal{D}$. Reward after 25 search iterations saturates beyond 1,332 demonstrations.}
\label{tab:distill_ablations}
\end{table}

Table~\ref{tab:distill_ablations} (left) compares two policy distillation targets. Training the policy on all feasible edges gives $52.8\%$ success in subsequent search, compared to $32.3\%$ when training only on the edges of successful plans, which omit other useful high-progress partial plans in the tree (Section~\ref{app:policy_target}). Table~\ref{tab:distill_ablations} (right) re-pretrains the initial policy on fractions of the sub-task dataset $\mathcal{D}$ and runs 25 search iterations with it. Reward saturates beyond $1{,}332$ demonstrations, indicating that our sub-task dataset is already sufficient.

\subsection{Does MCTS converge?}
\label{app:mcts_convergence}

\begin{figure}[t]
    \centering
    \includegraphics[width=0.8\linewidth]{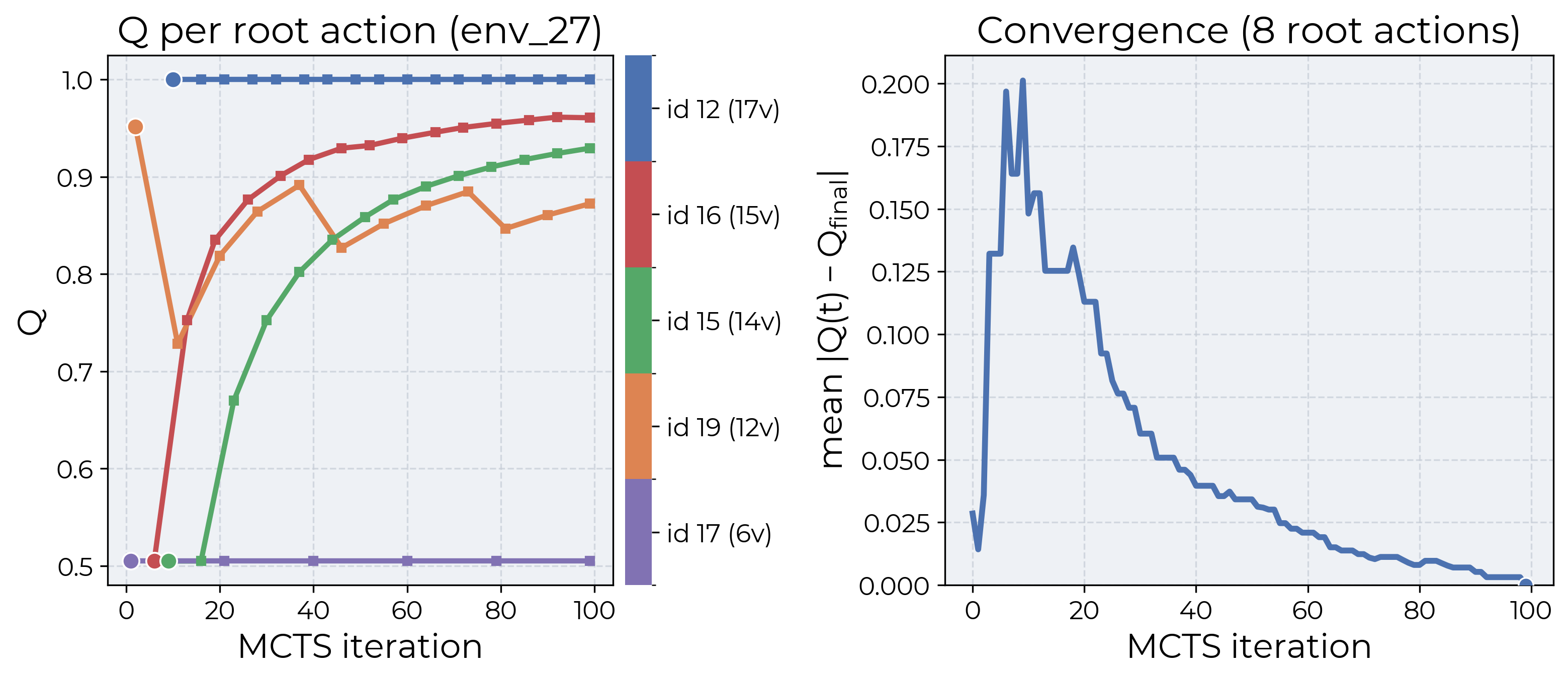}
    \caption{\small\textbf{Convergence of $Q$ values.} We visualize the convergence of root-action value estimates during MCTS in a representative training environment. Left: $Q$ trajectories for representative depth-1 actions, where markers indicate updates on iterations when the corresponding action is selected. Right: mean absolute gap to the final estimate, $\lvert Q(t)-Q_{\mathrm{final}}\rvert$, averaged over all visited root actions. Root-action values fan out from noisy early estimates and converge steadily as search budget increases.}
    \label{fig:mcts_convergence}
\end{figure}

Figure~\ref{fig:mcts_convergence} visualizes how MCTS refines the value estimates of root actions during search. Each depth-1 node stores a running estimate updated by incremental averaging over rollout returns, so its $Q$ value changes only on iterations where that action is selected. The left panel shows representative root-action trajectories in a single training environment. Some actions begin with pessimistic single-rollout estimates and then increase gradually as repeated visits discover successful downstream subtrees, while clearly worse actions are revisited only a few times and then abandoned. The most reliable action remains at high value throughout and receives the largest share of visits. The right panel summarizes this behavior over all visited root actions by plotting the mean gap to the final estimate, $\lvert Q(t)-Q_{\mathrm{final}}\rvert$. The gap decreases rapidly early in search and then decays more slowly, indicating that MCTS identifies the relative quality of root actions early and continues refining their values with additional budget. This is important for DISCO because it shows that the tree search is not only exploring actions, but also progressively producing more stable value supervision.

\subsection{Do better nodes get visited more?}
\label{app:q_vs_visits}

\begin{figure}[t]
    \centering
    \includegraphics[width=0.6\linewidth]{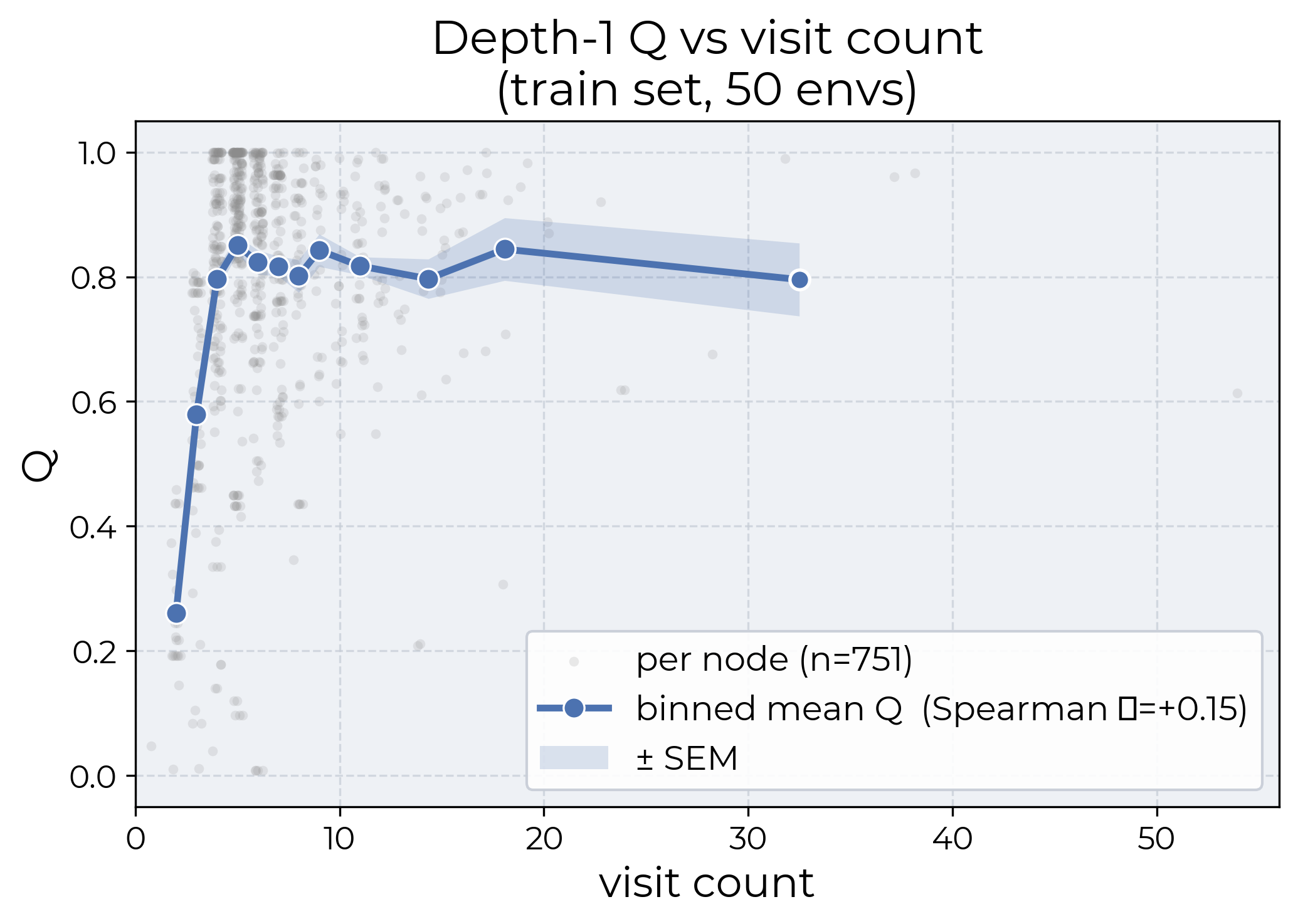}
    \caption{\small\textbf{Visit count compared to $Q$-values.} We plot the MCTS value estimate of each visited root action against its visit count across the training trees. Gray points show individual depth-1 nodes, and the blue curve shows the binned mean with standard error. Spearman $\rho$ measures rank correlation between visit count and $Q$ without assuming a linear relationship. The positive but modest correlation ($\rho=+0.15$) reflects an early rise followed by a plateau. Mean $Q$ rises sharply over the first few visits and then saturates, indicating that MCTS quickly abandons poor actions and allocates additional search budget primarily to actions already confirmed to be promising.}
    \label{fig:q_vs_visits}
\end{figure}

Figure~\ref{fig:q_vs_visits} compares root-action visit count to the corresponding MCTS value estimate across the training trees. Each point is a visited depth-1 node, and the blue curve shows the binned mean with standard error. The main trend is two-stage. For very low visit counts, mean $Q$ increases sharply with the number of visits, indicating that actions visited only a few times are often those that MCTS quickly identifies as poor and stops exploring. Beyond roughly five visits, the mean value plateaus near a high-$Q$ regime, showing that actions receiving sustained search budget are already consistently promising. The Spearman correlation reports the rank correlation between visit count and $Q$, measuring whether higher-visit actions tend to have higher value without assuming a linear relationship. The correlation is positive but modest ($\rho=+0.15$), which matches the shape of the plot: the main effect is an early jump followed by a plateau, not a linear trend. This indicates that visit allocation tracks action quality, but also that visit count alone becomes a weak proxy for ranking actions after early pruning. This motivates the use of a learned value function for finer action ranking during rollout.

\subsection{Why Progressive Widening Matters for Distillation}
\label{app:pw_analysis}

\begin{figure}[t]
\centering
\includegraphics[width=\linewidth]{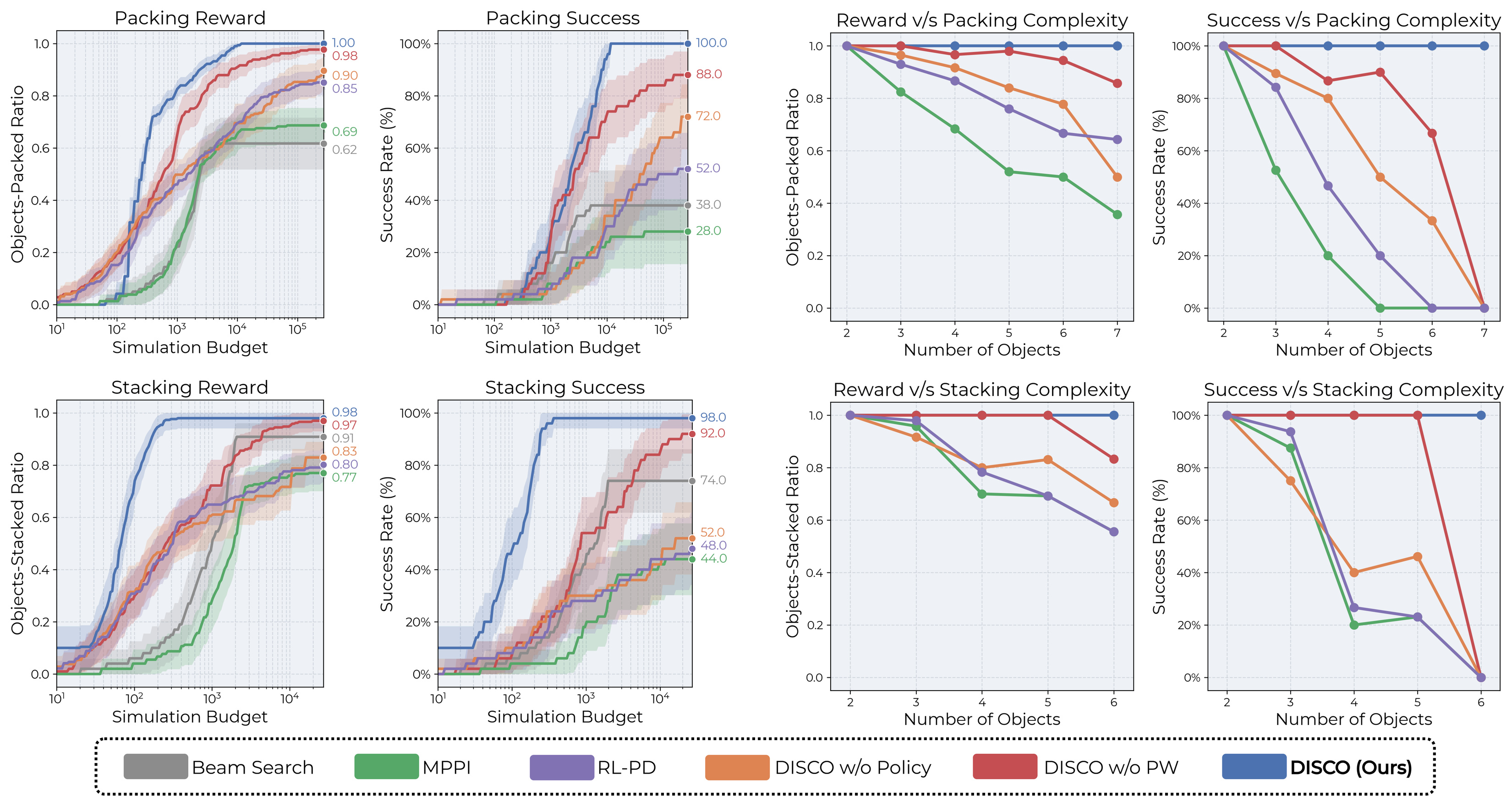}
\caption{\small \textbf{Simulation Results.} DISCO improves reward and success with increasing simulation budget on shelf packing and object stacking, and remains more robust as task complexity increases.}
\label{fig:results2}
\end{figure}

We re-display our main results in Figure~\ref{fig:results2} and Table~\ref{tab:results_main2} for ease of viewing.

\begin{table}[t]
\centering
\small
\resizebox{\textwidth}{!}{%
\begin{tabular}{l|ccc|ccc}
\toprule
& \multicolumn{3}{c|}{\textbf{Shelf Packing (Sim) Metrics}}
& \multicolumn{3}{c}{\textbf{Object Stacking Metrics}} \\[5px]
Algorithm
& Reward ($\uparrow$)
& Success (\%) ($\uparrow$)
& Train Plans ($\uparrow$)
& Reward ($\uparrow$)
& Success (\%) ($\uparrow$)
& Train Plans ($\uparrow$) \\
\midrule

Beam Search 
& \dd{24.3}{8.0} 
& \dd{6.7}{5.1} 
& \dd{4.26}{1.14}
& \dd{79.9}{4.2} 
& \dd{33.3}{13.6} 
& \dd{5.84}{1.00} \\

MPPI 
& \dd{36.7}{7.8} 
& \dd{13.3}{7.2} 
& \dd{12.20}{3.25}
& \dd{60.7}{6.2} 
& \dd{13.3}{12.2} 
& \dd{3.32}{0.50} \\

RL-PD 
& \dd{71.0}{4.8} 
& \dd{20.0}{13.7} 
& \dd{138.52}{40.81}
& \dd{64.5}{5.7} 
& \dd{20.0}{8.5} 
& \ddbf{46.46}{10.13} \\

DISCO w/o Policy 
& \dd{65.9}{6.2} 
& \dd{6.7}{2.3} 
& \dd{7.56}{1.59}
& \dd{75.2}{5.0} 
& \dd{33.3}{8.9} 
& \dd{3.02}{0.56} \\

DISCO w/o PW 
& \dd{39.6}{7.2} 
& \dd{0.0}{0.0} 
& \dd{133.14}{31.88}
& \dd{64.8}{7.5} 
& \dd{26.7}{10.8} 
& \dd{22.98}{3.79} \\

\cellcolor{tablecolor}\textbf{DISCO (Ours)} 
& \ddbf{90.9}{6.7} 
& \ddbf{80.0}{9.5} 
& \ddbf{463.16}{79.76}
& \ddbf{94.7}{2.0} 
& \ddbf{66.7}{5.5} 
& \dd{34.12}{5.78} \\

\bottomrule
\end{tabular}}
\vspace{5pt}
\caption{\small \textbf{Simulation Metrics}. We report average reward and success rate for DISCO against baselines on a held-out test set for shelf packing and object stacking. We also report number of plans found by each method during training. Results are averaged over 32 environment seeds and show a 95\% confidence interval.}
\label{tab:results_main2}
\end{table}

Table~\ref{tab:results_main2} and Figure~\ref{fig:results2} show that progressive widening affects not only search-time coverage, but also the supervision seen by the policy-value model. With a fixed child set, the value function observes a narrower set of alternatives from each partial state. With progressive widening, the value function receives supervision over a broader set of competing futures, which better matches its role at test time: selecting among multiple plausible actions proposed by the policy. Thus, progressive widening contributes both to finding plans and to producing more informative policy-value supervision.

\subsection{Interpretation of Baseline Behavior}
\label{app:baseline_interpretation}

Table~\ref{tab:results_main2} and Figure~\ref{fig:results2} show that MPPI performs policy-guided continuous optimization but does not maintain a branching tree over alternative futures, making it harder to recover from locally promising but globally poor sequences. RL-PD and the DISCO ablations can achieve competitive reward in some regimes, indicating that learned policies, online interaction, or MCTS structure can all produce useful partial behavior. However, their success drops faster as object count increases, suggesting that partial progress alone is insufficient for long-horizon completion. DISCO remains more stable because the policy prior reduces search width, the value function reduces effective search depth, and progressive widening continues expanding search into new continuous action regions as more budget becomes available.

\subsection{Search as Training-Time Supervision}
\label{app:search_as_supervision}

Table~\ref{tab:results_main2} and Figure~\ref{fig:results2} show that DISCO's gains are not only due to spending more computation during deployment. MCTS is used during training to generate long-horizon supervision, and the final policy-value model inherits enough of this structure to solve new tasks through direct rollout. The train-plan counts support this view: in Shelf Packing, DISCO finds substantially more successful plans than the baselines, giving the distillation stage more long-horizon examples. However, plan count alone does not fully explain performance, since methods can find nontrivial numbers of plans without matching DISCO's held-out success.

\section{Additional Qualitative Analysis}
\label{app:qualitative}

In this section, we visualize graphs generated during DISCO's distillation process to analyze how DISCO operates in key decision states and funnel states (where only one action leads to success). We also analyze the nodes that progressive widening finds later during MCTS search, and how they affect performance. We also show multi-modality in action proposal and planning.

\subsection{Analysis of key decision states}

\begin{figure}[t]
    \centering
    \includegraphics[width=\linewidth]{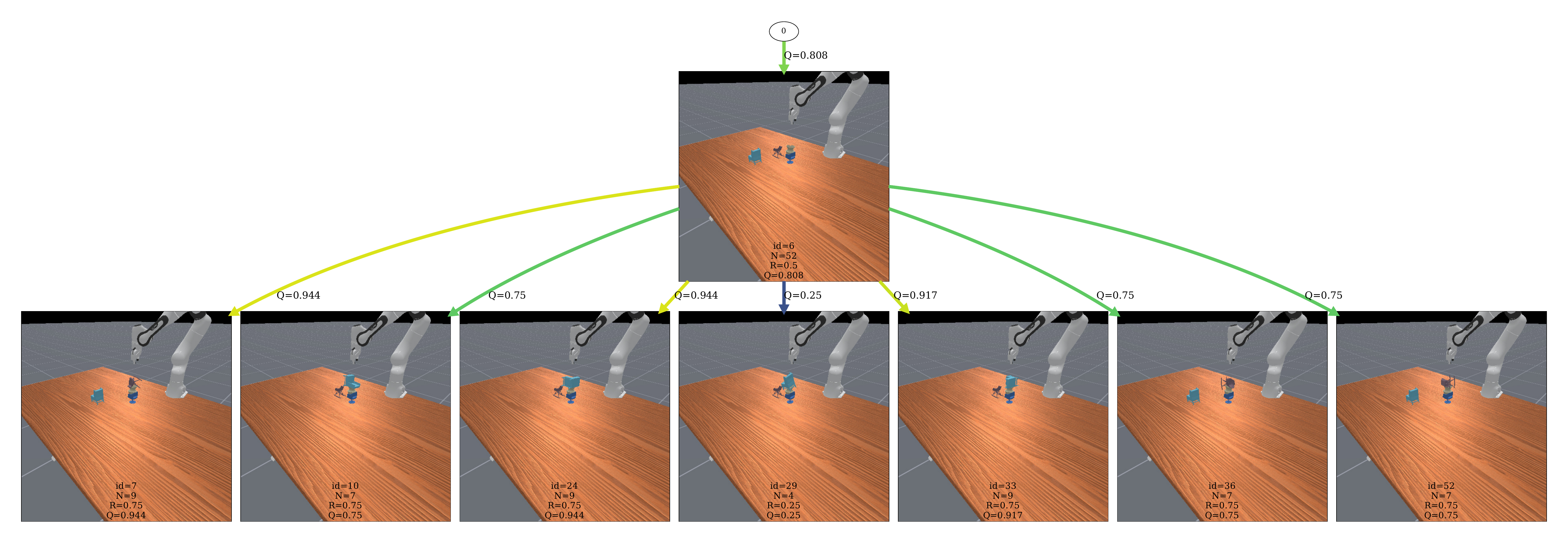}
    \caption{\small\textbf{Decision Making}. The graph above shows an object stacking task vertex with several children. In this graph we see that the higher Q-values of 0.944 and 0.917 are given most often to stable placements of the light blue object, by placing it on flat surfaces, rather than the unstable placement in node id=29 with a Q-value of 0.25}
    \label{fig:decision_1}
\end{figure}

Figure~\ref{fig:decision_1} shows a visualization of DISCO choosing actions from a vertex in the tree.

\subsection{Analysis of funnel states}

Figure~\ref{fig:funnel_1} in the main paper shows a funnel state in an object stacking DISCO tree, where only one of the proposed actions leads to success. At this vertex (id=5), DISCO strongly prefers placing the pink mug before the blue object, since the mug provides stable support for the blue object to be placed later, and the Q-values reflect this preference.

\subsection{Progressive widening can find better actions later}

\begin{figure}[t]
    \centering
    \includegraphics[width=\linewidth]{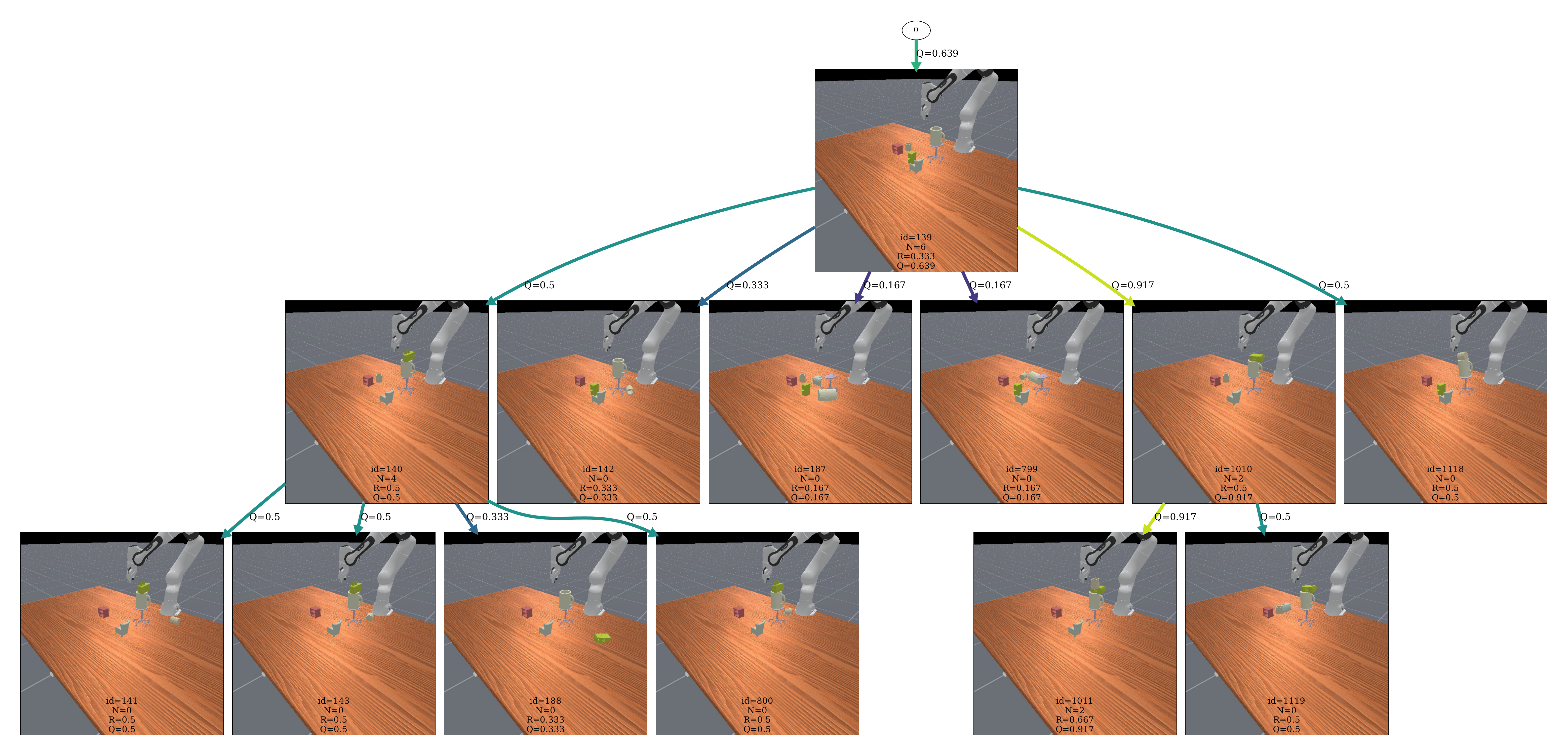}
    \caption{\small\textbf{Mind Change}. The object stacking graph above shows how the DISCO algorithm changes its mind after progressive widening adds a new action and a new vertex. Here, vertex id=140 was the preferred vertex with more objects getting stacked. Later during search, vertex id=1010 is added, offering a better placement for the yellow object, that allows more objects to be placed on top. Thus, progressive widening leads DISCO to find better actions later during search.}
    \label{fig:mind_change_1}
\end{figure}

Figure~\ref{fig:mind_change_1} shows how DISCO can find successful plans as progressive widening adds more vertices to the graph.

\subsection{Multi-modality}

\begin{figure}[t]
    \centering
    \includegraphics[width=\linewidth]{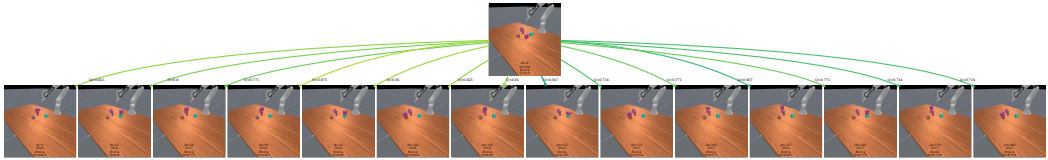}
    \caption{\small\textbf{Multi-Modal Proposals}. The graph above shows multi-modal action proposals of object stacking originating from a single node.}
    \label{fig:multi_modal_proposal_1}
\end{figure}

\begin{figure}[t]
    \centering
    \includegraphics[width=\linewidth]{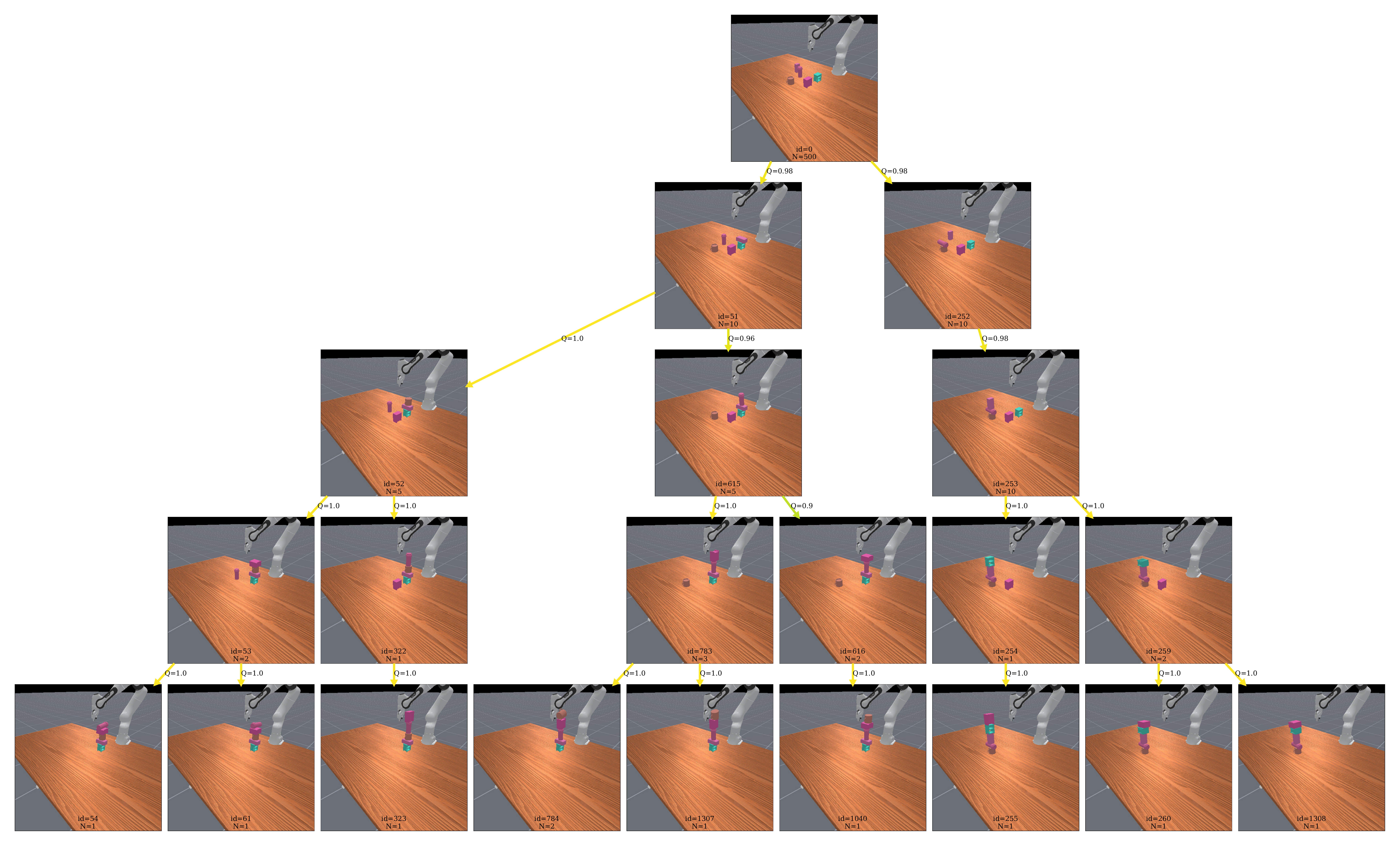}
    \caption{\small\textbf{Multi-Modal Plans}. The graph of above visualizes multi-modal plans found by DISCO. Here, we show a graph of the object stacking environment where DISCO finds many different ways of creating the stack of objects.}
    \label{fig:multi_modal_plan_1}
\end{figure}

Figure~\ref{fig:multi_modal_proposal_1} shows multi-modal action proposals with DISCO's policy and Figure~\ref{fig:multi_modal_plan_1} shows how DISCO can find multiple diverse plans for solving the same object stacking task.

\clearpage 
\section{Mathematical Details}
\label{app:derivations}

This section formalizes the problem statement of Section~\ref{sec:problem} and gives the reasoning behind the design choices of Section~\ref{sec:method}. Section~\ref{app:markov} casts each task instance as a Markov decision process. Section~\ref{app:success_probability_objective} states the objective and the success probability that $Q_\phi$ estimates. Section~\ref{app:policy_target} motivates the policy distillation target, and Section~\ref{app:value_prior} explains how the progress reward and $Q_\phi$ interact during search.

\subsection{Markovian Formulation}
\label{app:markov}
\label{app:policy_plan_distribution}

\textbf{Task instances as Markov decision processes.} Each task instance $\langle s_0,\mathcal{G}\rangle\sim\mathcal{T}$ defines a deterministic, finite-horizon, goal-conditioned Markov decision process $\langle\mathcal{S},\mathcal{A},f_{\mathrm{sim}},\mathcal{G},K\rangle$. The state $s\in\mathcal{S}$ collects the 6D poses of all objects. An action $\mathbf{a}\in\mathcal{A}$ specifies a grasp pose and a place pose for one object. The motion planner~\cite{curobo} converts $\mathbf{a}$ into a joint trajectory, which the simulator executes, and we fold both steps into a single transition model $f_{\mathrm{sim}}$:
\begin{equation}
\label{eq:app_transition}
    s_{k+1}=f_{\mathrm{sim}}(s_k,\mathbf{a}_k).
\end{equation}
The Markov property says that the outcome of an action depends only on the current object poses and the action, and not on how the objects reached these poses. An action is feasible in state $s$ if motion planning succeeds and the executed motion keeps the scene valid. We denote the set of feasible actions by $\mathcal{F}(s)\subseteq\mathcal{A}$. An episode terminates at step $T\leq K$ when the goal is reached ($\mathcal{G}(s_T)=1$), when the horizon $K$ is exhausted, or when no feasible action remains ($\mathcal{F}(s_T)=\emptyset$). Task success is a sparse terminal outcome $\mathbbm{1}_{\mathcal{G}(s_T)}$. The progress reward $r$ of Section~\ref{sec:problem} is used only to guide search during training (Section~\ref{app:value_prior}).

\textbf{Policy-induced plan distribution.} For a task instance with initial state $s_0$, a plan is $\rho=[s_0,\mathbf{a}_0,s_1,\ldots,\mathbf{a}_{T-1},s_T]$. Under the Markov assumption, the plan distribution factorizes as
\begin{equation}
\label{eq:app_plan_factorization}
    p(\rho\mid s_0)
    =
    \prod_{k=0}^{T-1}
    p(s_{k+1}\mid s_k,\mathbf{a}_k)\,
    p(\mathbf{a}_k\mid s_k).
\end{equation}
The robot does not choose actions from the state $s_k$ directly, but from the segmented point cloud $\mathbf{o}_k$ observed in that state. We therefore factor action selection through the observation:
\begin{equation}
\label{eq:app_observation_factorization}
    p(\mathbf{a}_k\mid s_k)
    =
    \int_{\mathbf{o}_k}
    p(\mathbf{a}_k\mid\mathbf{o}_k)\,
    p(\mathbf{o}_k\mid s_k)
    \,d\mathbf{o}_k.
\end{equation}
In simulation, the observation function and the transition are deterministic:
\begin{equation}
    \mathbf{o}_k=\Omega(s_k),
    \qquad
    s_{k+1}=f_{\mathrm{sim}}(s_k,\mathbf{a}_k).
\end{equation}
Thus, the only stochastic component is the action distribution, which we model with the policy,
\begin{equation}
\begin{gathered}
    p(\rho\mid s_0)
    =
    \prod_{k=0}^{T-1}
    p(s_{k+1}\mid s_k,\mathbf{a}_k)\,
    p(\mathbf{a}_k\mid s_k), \\
    p(\mathbf{a}_k\mid s_k)
    \approx
    \pi_\theta(\mathbf{a}_k\mid\mathbf{o}_k).
\end{gathered}
\end{equation}
This gives the policy-induced plan distribution
\begin{equation}
\label{eq:app_policy_induced_plan}
    p_\theta(\rho\mid s_0)
    =
    \prod_{k=0}^{T-1}
    \pi_\theta(\mathbf{a}_k\mid\mathbf{o}_k)
\end{equation}

\subsection{Success Probability Objective}
\label{app:success_probability_objective}

\textbf{Objective.} The probability of solving the task from $s_0$ can be written as an expectation over policy-induced plans:
\begin{equation}
\label{eq:app_success_probability}
\begin{aligned}
    p_\theta(\mathcal{G}(s_T)=1\mid s_0)
    &=
    \int_{\rho}
    p(\mathcal{G}(s_T)=1\mid \rho,s_0)\,
    p_\theta(\rho\mid s_0)\,d\rho \\
    &=
    \mathbb{E}_{\rho\sim p_\theta(\rho\mid s_0)}
    \left[
    \mathbbm{1}_{\mathcal{G}(s_T)}
    \right],
\end{aligned}
\end{equation}
since the plan determines its final state $s_T$. The objective of Section~\ref{sec:problem} is to maximize this probability of reaching the goal over task instances:
\begin{equation}
\label{eq:app_theta_objective}
    \theta^*
    =
    \arg\max_\theta\;
    \mathbb{E}_{\langle s_0,\mathcal{G}\rangle\sim\mathcal{T}}\;
    \mathbb{E}_{\rho\sim p_\theta(\rho\mid s_0)}
    \left[
    \mathbbm{1}_{\mathcal{G}(s_T)}
    \right].
\end{equation}
During training, we only have access to the simulated instances $\mathcal{T}_{\mathrm{sim}}$, and generalization to unseen instances comes from conditioning the networks on point clouds. Direct optimization of Eq.~\ref{eq:app_theta_objective} is difficult because success is sparse and only observed at the end of long sequences of continuous actions. DISCO therefore uses MCTS in simulation to find successful plans and distills the resulting search trees into $\pi_\theta$ and $Q_\phi$.

\textbf{Success probability.} For a policy $\pi$, the probability of reaching the goal after executing action $\mathbf{a}$ in state $s$ and following $\pi$ afterwards is
\begin{equation}
\label{eq:app_success_q}
    Q^{\pi}(s,\mathbf{a})
    =
    P\big(\mathcal{G}(s_T)=1 \mid s_k=s,\ \mathbf{a}_k=\mathbf{a},\ \pi\big),
\end{equation}
and the success probability from $s$ is $\mathbb{E}_{\mathbf{a}\sim\pi}[Q^{\pi}(s,\mathbf{a})]$. The target $y_i$ in Eq.~\ref{eq:value_distill_loss} is a Monte-Carlo estimate of it: each visited leaf below $\mathbf{v}_i$ ends one continuation after executing $\mathbf{a}_i$ from the parent state, so $y_i$ is the empirical success rate over these continuations. Since the continuations follow the search rather than the raw policy, $y_i$ estimates the success probability achievable with search, which is the improvement we want to distill. The empirical rate over $n=|\mathrm{leaves}(\mathbf{v}_i)|$ continuations has variance of roughly $y_i(1-y_i)/n$, so targets backed by more leaves are more reliable. The weight $w_i=\log(1+n)$ grows with $n$, but sublinearly, so that large subtrees do not dominate the loss.

\textbf{Action selection.} During value-guided rollouts and at test time, we sample $B$ actions from $\pi_\theta$ and execute $\mathbf{a}^{*}=\arg\max_b Q_\phi(\mathbf{o},\mathbf{a}^{(b)})$. If $Q_\phi$ were exact, then
\begin{equation}
\label{eq:app_best_of_b}
    Q^{\pi_\theta}(s,\mathbf{a}^{*})
    =
    \max_{b} Q^{\pi_\theta}(s,\mathbf{a}^{(b)})
    \;\geq\;
    \mathbb{E}_{\mathbf{a}\sim\pi_\theta}\big[Q^{\pi_\theta}(s,\mathbf{a})\big],
\end{equation}
in expectation over the samples, so by the policy improvement theorem~\cite{suttonbarto}, acting with this best-of-$B$ rule succeeds at least as often as sampling from $\pi_\theta$. The policy therefore only needs to place probability mass on good actions, and $Q_\phi$ selects among them.

\subsection{Policy Distillation Target}
\label{app:policy_target}

At an expanded vertex with state $s$ and observation $\mathbf{o}$, search samples candidates $\mathbf{a}^{(b)}\sim\pi_\theta(\mathbf{a}\mid\mathbf{o})$ and keeps only those in $\mathcal{F}(s)$. Training the policy on all feasible edges therefore fits, at each expanded state, the distribution
\begin{equation}
\label{eq:app_policy_target}
    \pi_{\mathrm{new}}(\mathbf{a}\mid\mathbf{o})
    \;\propto\;
    \pi_\theta(\mathbf{a}\mid\mathbf{o})\,
    \mathbbm{1}\big[\mathbf{a}\in\mathcal{F}(s)\big].
\end{equation}
This has two effects. First, it removes probability mass from infeasible actions, so that more policy samples are executable in the next round of search. Second, the distribution of training states shifts toward long-horizon states. Progressive widening adds more children to more-visited vertices (Section~\ref{sec:mcts}), and visits concentrate on high-value subtrees, so states along promising plans contribute more training pairs. These states lie deep in long-horizon plans and are not covered by the single-step dataset $\mathcal{D}$.

An alternative is to train only on the edges of successful plans, which would fit $\pi_\theta$ restricted to actions that lead to success. Although this target is sharper, it discards feasible high-progress partial plans and yields far fewer training pairs. Table~\ref{tab:distill_ablations} (left) shows that training on all feasible edges gives higher success in subsequent search. This division of labor is consistent with Eq.~\ref{eq:app_best_of_b}: the policy proposes a diverse set of feasible actions, and $Q_\phi$ ranks them by success probability.

\subsection{Value Prior and Progress Backups}
\label{app:value_prior}

During search, each vertex value $Q(\mathbf{v})$ is a running average of the cumulative progress return $R=\sum_{\mathbf{v}_j\in\mathcal{P}_T} r_j$ along the path $\mathcal{P}_T$ from the root to the leaf of each descent. From the second distillation round onward, the value of a newly added, unvisited child is initialized with $Q_\phi(\mathbf{o}_p,\mathbf{a})$, an estimate of success probability. On its first visit, this prior is replaced by the backed-up return. Unvisited children are therefore scored by $Q_\phi$, while visited children are scored by progress returns.

These two quantities are directly comparable. Since $r(s,\mathbf{a},s')=\kappa(s')-\kappa(s)$, the return telescopes to $R=\kappa(s_T)-\kappa(s_0)$. Task progress lies in $[0,1]$ and $\kappa(s_0)=0$ at the root in our simulated tasks, so $R=\kappa(s_T)\in[0,1]$, and a leaf satisfies the goal exactly when its progress reaches the maximum value, i.e., $\mathcal{G}(s_T)=1 \Leftrightarrow \kappa(s_T)=1 \Leftrightarrow R=1$. For any distribution over continuations below a child, Markov's inequality then gives
\begin{equation}
\label{eq:app_markov_prior}
    P\big(\mathcal{G}(s_T)=1\big) = P(R\geq 1) \leq \mathbb{E}[R].
\end{equation}
The success probability estimated by $Q_\phi$ is thus a lower bound on the expected progress return that would be backed up into the same child. Hence $Q_\phi$ acts as a conservative prior: it never overstates an unvisited child relative to the progress returns of its visited siblings, while the UCB exploration term, which is largest for unvisited children ($n_{\mathrm{edge}}+1=1$), still ensures that they are explored.

Finally, the progress reward $r$ only shapes the search statistics used to select vertices during training. The value target $y_i$ in Eq.~\ref{eq:value_distill_loss} is computed solely from goal indicators $\mathbbm{1}_{\mathcal{G}(s_\ell)}$ of the leaves, and never uses $r$. Consequently, $Q_\phi$ is trained to estimate success probability alone, and only $Q_\phi$, not the progress reward, is used to select actions at test time.

\end{document}